\def\swordtitleruleplacement{1}
\def\swordtitlebrandvariant{2}
\def\swordtitlefontvariant{1}
\documentclass[oneside,a4paper,onecolumn,10pt]{article}
\usepackage{array}
\usepackage{tabularx}
\usepackage{longtable}
\usepackage[dvipsnames]{xcolor}

\definecolor{linkblue}{RGB}{0,82,155}
\definecolor{citegreen}{RGB}{0,110,70}
\definecolor{urlviolet}{RGB}{100,40,130}

\definecolor{swordcoral}{RGB}{194,64,47}
\definecolor{swordblue}{RGB}{58,84,176}
\definecolor{swordviolet}{RGB}{105,74,158}

\definecolor{lstbg}{RGB}{250,246,239}
\definecolor{lstframe}{RGB}{223,220,213}
\definecolor{lstcmt}{RGB}{0,110,70}
\definecolor{lstrole}{RGB}{0,82,155}
\definecolor{lststr}{RGB}{60,60,60}
\definecolor{lstkw}{RGB}{140,30,120}
\definecolor{hlcol}{RGB}{250,246,239}

\definecolor{tokfg}{HTML}{374151}
\definecolor{tokbg}{HTML}{F3F4F6}

\definecolor{bordo}{RGB}{128,0,32}

\definecolor{panelPrompt}{HTML}{FCF3EA}
\definecolor{panelReason}{HTML}{F4F4F4}
\definecolor{panelAnswer}{HTML}{E8E8E4}
\definecolor{panelBorder}{HTML}{B9B9B9}
\definecolor{panelInk}{HTML}{27272A}
\definecolor{panelInkStrong}{HTML}{141414}
\definecolor{panelAccent}{HTML}{C43C45}
\definecolor{panelAccentBG}{HTML}{F3DEDE}

\usepackage{fullpage}
\usepackage[left=2.5cm,top=1.7cm,bottom=2.2cm,right=2.5cm]{geometry}
\usepackage[utf8]{inputenc}
\usepackage[T2A,T1]{fontenc}
\usepackage{iftex}
\ifPDFTeX
  \usepackage{CJKutf8}
  \DeclareUnicodeCharacter{211D}{\ensuremath{\mathbb{R}}}
  \DeclareUnicodeCharacter{2202}{\ensuremath{\partial}}
  \DeclareUnicodeCharacter{2212}{\ensuremath{-}}
\fi

\usepackage{amsmath}
\usepackage{amssymb}
\usepackage{bbm}
\usepackage{bm}

\usepackage{booktabs}
\usepackage{multirow}
\usepackage{colortbl}

\usepackage{graphicx}
\usepackage{wrapfig}
\usepackage{needspace}
\usepackage{float}
\usepackage[most]{tcolorbox}

\usepackage{pifont}
\usepackage{fontawesome5}
\usepackage{soul}
\usepackage{eurosym}
\usepackage{xspace}
\usepackage{seqsplit}
\usepackage{comment}
\usepackage{enumitem}

\usepackage{tgpagella}
\usepackage{roboto}
\usepackage{nopageno}

\usepackage[authoryear,round]{natbib}

\usepackage[
  colorlinks=true,
  linkcolor=swordblue,
  citecolor=swordblue,
  urlcolor=swordblue,
  unicode
]{hyperref}
\usepackage{xurl}
\usepackage{footnotehyper}
\makesavenoteenv{figure}

\usepackage{etoc}
\usepackage[nameinlink,capitalise,noabbrev]{cleveref}

\usepackage{siunitx}

\usepackage{tikz}
\usetikzlibrary{positioning, arrows.meta, calc, decorations.pathreplacing, fit, backgrounds, patterns}
\usepackage{pgfplots}
\usepgfplotslibrary{groupplots}
\pgfplotsset{compat=1.16}
\usepackage{fancyvrb}
\usepackage{fvextra}
\usepackage{algorithm}
\usepackage{algpseudocode}
\usepackage{listings}
\usepackage{CJKutf8}

\usepackage{iftex}
\ifPDFTeX
  \newcommand{\bodyfont}{\fontfamily{XCharter-TLF}\selectfont}
  \newcommand{\labelfont}{\fontfamily{phv}\selectfont}
  \newcommand{\paperromanfont}{\normalfont}
\else
  \usepackage{fontspec}
  \newfontfamily\bodyfont{Spectral}[
    Path=fonts/, Extension=.ttf,
    UprightFont=*-Regular, ItalicFont=*-Italic,
    BoldFont=*-Bold, BoldItalicFont=*-BoldItalic]
  \newfontfamily\labelfont{LibreFranklin}[
    Path=fonts/, Extension=.ttf,
    UprightFont=*-SemiBold, BoldFont=*-Bold]
  \newfontfamily\paperromanfont{lmroman10-regular.otf}[Ligatures=TeX]
\fi

\ifPDFTeX
  \newcommand{\fg}[2][gbsn]{%
  \begin{CJK}{UTF8}{#1}#2\end{CJK}%
}
\else
  \newfontfamily\intlfont{NotoSansMonoCJKsc-Regular.otf}[Path=fonts/, Scale=MatchLowercase]
  \newcommand{\fg}[1]{{\intlfont #1}}
  \IfFontExistsTF{Noto Sans Devanagari}
    {\newfontfamily\devfont{Noto Sans Devanagari}[Scale=MatchLowercase]}{\let\devfont\intlfont}
  \IfFontExistsTF{Noto Sans Georgian}
    {\newfontfamily\geofont{Noto Sans Georgian}[Scale=MatchLowercase]}{\let\geofont\intlfont}

\fi
\ifPDFTeX \fi

\lstdefinestyle{trace}{
  basicstyle=\ttfamily\scriptsize,
  backgroundcolor=\color{lstbg},
  frame=single, rulecolor=\color{lstframe}, framerule=0.6pt,
  breaklines=true, breakindent=0pt, postbreak=\mbox{\textcolor{lstframe}{$\hookrightarrow$}\space},
  columns=fullflexible, keepspaces=true,
  commentstyle=\color{lstcmt}\itshape,
  keywordstyle=\color{lstrole}\bfseries,
  emphstyle=\color{lstrole}\bfseries,
  morekeywords={messages,user,assistant,system,thinking,text,signature,primer,decoder,reconciler},
  morecomment=[l]{\#},
  aboveskip=2pt, belowskip=2pt,
  xleftmargin=5pt, xrightmargin=3pt, framexleftmargin=5pt,
}

\lstdefinestyle{pyclean}{
  language=Python,
  basicstyle=\ttfamily\scriptsize,
  backgroundcolor=\color{lstbg},
  frame=single, rulecolor=\color{lstframe}, framerule=0.6pt,
  breaklines=true, breakindent=0pt,
  postbreak=\mbox{\textcolor{lstframe}{$\hookrightarrow$}\space},
  columns=fullflexible, keepspaces=true,
  showstringspaces=false,
  commentstyle=\color{lstcmt}\itshape,
  keywordstyle=\color{lstkw}\bfseries,
  stringstyle=\color{lstrole},
  emphstyle=\color{lstrole}\bfseries,
  emph={client,messages,create,model,max_tokens,thinking,content,role,signature},
  aboveskip=6pt, belowskip=6pt,
  xleftmargin=5pt, xrightmargin=3pt, framexleftmargin=5pt,
}

\lstdefinestyle{extracttemplate}{
  basicstyle=\ttfamily\footnotesize,
  frame=none,
  backgroundcolor={},
  breaklines=true,
  breakindent=12pt,
  columns=fullflexible,
  keepspaces=true,
  showstringspaces=false,
  alsoletter={@},
  morekeywords={@thought},
  keywordstyle=\color{panelAccent}\bfseries,
  commentstyle=\color{panelAccent}\itshape,
  morecomment=[l]{//},
  xleftmargin=0pt,
  xrightmargin=0pt,
  aboveskip=0pt,
  belowskip=0pt,
}

\lstdefinestyle{extractcall}{
  language=Python,
  basicstyle=\ttfamily\scriptsize,
  frame=none,
  backgroundcolor={},
  breaklines=true,
  breakindent=12pt,
  columns=fullflexible,
  keepspaces=true,
  showstringspaces=false,
  commentstyle=\color{lstcmt}\itshape,
  keywordstyle=\color{lstkw}\bfseries,
  stringstyle=\color{lstrole},
  emphstyle=\color{panelAccent}\bfseries,
  emph={client,messages,create,generate_content,responses,models,Content,Part},
  xleftmargin=0pt,
  xrightmargin=0pt,
  aboveskip=0pt,
  belowskip=0pt,
}

\lstdefinestyle{judgeprompt}{
  basicstyle=\ttfamily\scriptsize,
  frame=none,
  backgroundcolor={},
  breaklines=true,
  breakindent=12pt,
  columns=fullflexible,
  keepspaces=true,
  showstringspaces=false,
  alsoletter={_-},
  emph={META,SUB,REAL,NOT-ARTIFACT,RESOLUTION,Reasoning trace},
  emphstyle=\color{panelAccent}\bfseries,
  xleftmargin=0pt,
  xrightmargin=0pt,
  aboveskip=0pt,
  belowskip=0pt,
}
\usepackage{xcolor}
\definecolor{accent}{HTML}{C43C45}

\usepackage[font=small,labelfont=bf,skip=4pt,format=plain]{caption}
\usepackage{fancyhdr}
\crefname{appendix}{Appendix}{Appendices}
\Crefname{appendix}{Appendix}{Appendices}
\crefname{app}{Appendix}{Appendices}
\Crefname{app}{Appendix}{Appendices}

\setitemize{itemsep=4pt,topsep=-3pt,parsep=0pt,partopsep=0pt,leftmargin=18pt}

\makeatletter
\renewcommand{\paragraph}{%
  \@startsection{paragraph}{4}%
  {\z@}{0.5ex \@plus 1ex \@minus .2ex}{-1em}%
  {\normalfont\normalsize\bfseries}%
}
\makeatother

\makeatletter
\let\origsection\section
\renewcommand\section{\@ifstar{\starsection}{\nostarsection}}
\newcommand\nostarsection[1]
{\sectionprelude\origsection{#1}\sectionpostlude}
\newcommand\starsection[1]
{\sectionprelude\origsection*{#1}\sectionpostlude}
\newcommand\sectionprelude{\vspace{-3mm}}
\newcommand\sectionpostlude{\vspace{-2.5mm}}
\makeatother

\makeatletter
\let\origsubsection\subsection
\renewcommand\subsection{\@ifstar{\starsubsection}{\nostarsubsection}}
\newcommand\nostarsubsection[1]
{\subsectionprelude\origsubsection{#1}\subsectionpostlude}
\newcommand\starsubsection[1]
{\subsectionprelude\origsubsection*{#1}\subsectionpostlude}
\newcommand\subsectionprelude{\vspace{-2mm}}
\newcommand\subsectionpostlude{\vspace{-2.5mm}}
\makeatother

\graphicspath{ {figures/} }

\sethlcolor{tokbg}

\tcbset{
  decodedpanel/.style={
    enhanced jigsaw,
    breakable,
    boxrule=0pt,
    arc=3.2pt,
    boxsep=0pt,
    left=6pt,
    right=6pt,
    top=5pt,
    bottom=6pt,
    before skip=6pt,
    after skip=6pt,
    coltext=panelInk,
  },
}

\newtcolorbox{decodedpromptpanel}{
  decodedpanel,
  colback=panelPrompt,
  fontupper=\sffamily\small\linespread{1.15}\selectfont,
}
\newtcolorbox{decodedreasoningpanel}{
  decodedpanel,
  colback=panelReason,
  fontupper=\ttfamily\scriptsize\linespread{1.15}\selectfont,
  before upper={\raggedright\sloppy},
}
\newtcolorbox{decodedanswerpanel}{
  decodedpanel,
  colback=panelAnswer,
  fontupper=\sffamily\small\linespread{1.15}\selectfont,
}

\newtcblisting{extractiontemplate}{
  decodedpanel,
  colback=panelPrompt,
  listing only,
  listing options={style=extracttemplate},
}
\newtcblisting{extractioncall}{
  decodedpanel,
  colback=panelReason,
  listing only,
  listing options={style=extractcall},
}
\newtcblisting{judgepromptpanel}{
  decodedpanel,
  colback=panelPrompt,
  listing only,
  listing options={style=judgeprompt},
}

\usepackage{arydshln}
\makeatletter
\def\adl@drawiv#1#2#3{%
        \hskip.5\tabcolsep
        \xleaders#3{#2.5\@tempdimb #1{1}#2.5\@tempdimb}%
                #2\z@ plus1fil minus1fil\relax
        \hskip.5\tabcolsep}
\newcommand{\cdashlinelr}[1]{%
  \noalign{\vskip 2pt
           \global\let\@dashdrawstore\adl@draw
           \global\let\adl@draw\adl@drawiv}
  \cdashline{#1}[.4pt/2pt]
  \noalign{\global\let\adl@draw\@dashdrawstore
           \vskip 2pt}}
\makeatother

\usepackage{subcaption}
\usepackage{microtype}
\definecolor{BoardBorder}{HTML}{E3E7EE}
\definecolor{BoardInk}{HTML}{1B2233}
\newtcblisting{promptboard}[1]{enhanced,breakable,listing only,colback=white,colframe=BoardBorder,
  boxrule=0.45pt,arc=1mm,boxsep=0pt,left=9pt,right=9pt,top=7pt,bottom=7pt,
  lefttitle=9pt,righttitle=9pt,toptitle=5pt,bottomtitle=5pt,
  colbacktitle=BoardBorder!35!white,coltitle=BoardInk,fonttitle=\bfseries\fontsize{9.5}{12}\selectfont,title={#1},
  before skip=10pt,after skip=10pt,
  listing options={basicstyle=\ttfamily\footnotesize,breaklines=true,breakindent=0pt,columns=fullflexible,keepspaces=true,showstringspaces=false}}

\newcommand{\methodname}{\textsc{Dynalize}}

\definecolor{OursRow}{HTML}{FDEDF0}
\newcommand{\gain}[1]{\textsubscript{\textcolor{green!45!black}{#1}}}
\newcommand{\gainlight}[1]{\textsubscript{\textcolor{green!45!black!45}{#1}}}

\title{Mitigating Reward Hacking in Preference Reward Models\\with Evolving Penalties}
\author{%
{\normalsize
\href{https://openreview.net/profile?id=~Afonso_Baldo1}{\textcolor{black}{\textbf{Afonso Baldo}}}\textsuperscript{1,\,3},\enspace
\href{https://openreview.net/profile?id=~Ant\%C3\%B3nio_Farinhas1}{\textcolor{black}{\textbf{António Farinhas}}}\textsuperscript{1},\enspace
\href{https://openreview.net/profile?id=~Marcos_Vinicius_Treviso1}{\textcolor{black}{\textbf{Marcos Treviso}}}\textsuperscript{2,\,3,\,4,\,5},\enspace
\href{https://openreview.net/profile?id=~Ricardo_Rei1}{\textcolor{black}{\textbf{Ricardo Rei}}}\textsuperscript{1},\enspace
\href{https://openreview.net/profile?id=~Nuno_M_Guerreiro1}{\textcolor{black}{\textbf{Nuno M. Guerreiro}}}\textsuperscript{1}}\\[1.1mm]
{\small \textsuperscript{1}Sword Health \qquad \textsuperscript{2}Gandara AI \qquad \textsuperscript{3}Instituto Superior Técnico, Universidade de Lisboa}\\[0.45mm]
{\small \textsuperscript{4}Instituto de Telecomunicações \qquad \textsuperscript{5}ELLIS Unit Lisbon}\\[0.45mm]
{\small \href{mailto:ai.research@sword.com}{\textcolor{black}{{\footnotesize\faEnvelope}\hspace{0.45em}ai.research@sword.com}}}
}
\date{}

\providecommand{\swordbodyfontvariant}{0}
\ifnum\swordbodyfontvariant=1\relax
  \let\paperromanfont\rmfamily
\fi

\providecommand{\swordtitlefontvariant}{0}
\ifnum\swordtitlefontvariant=1\relax
  \ifPDFTeX
    \newcommand{\swordtitleface}{\fontfamily{XCharter-TLF}\selectfont}
  \else
    \newfontfamily\swordtitleface{XCharter}[
      UprightFont    = XCharter-Roman,
      ItalicFont     = XCharter-Italic,
      BoldFont       = XCharter-Bold,
      BoldItalicFont = XCharter-BoldItalic
    ]
  \fi
\else
  \newcommand{\swordtitleface}{}
\fi

\providecommand{\swordtitleruleplacement}{1}
\providecommand{\swordtitlebrandvariant}{2}
\newcommand{\swordtitlelogo}{%
  \ifnum\swordtitlebrandvariant=1\relax
    \includegraphics[width=2.25cm]{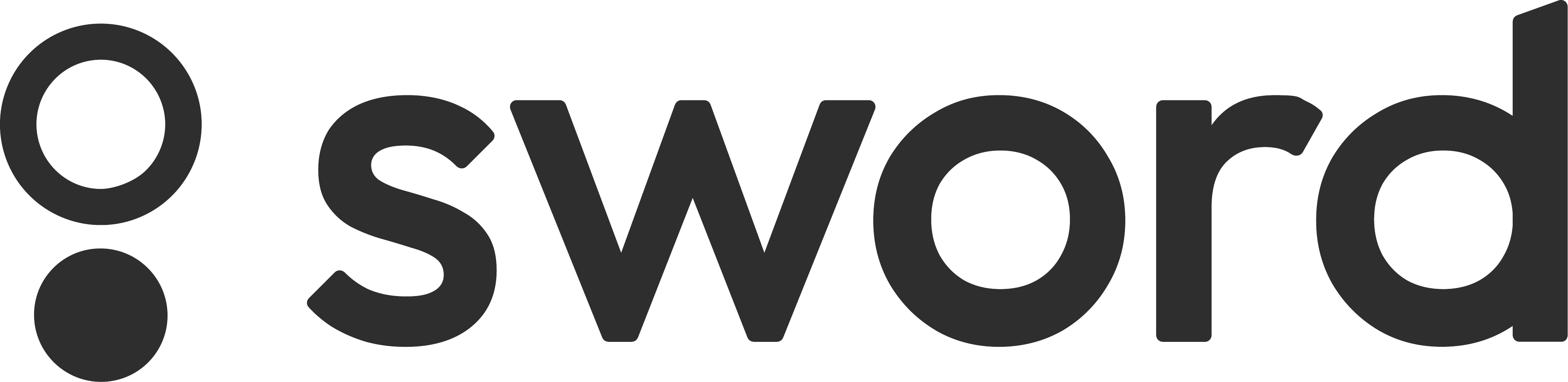}%
  \else
    \includegraphics[width=2.25cm]{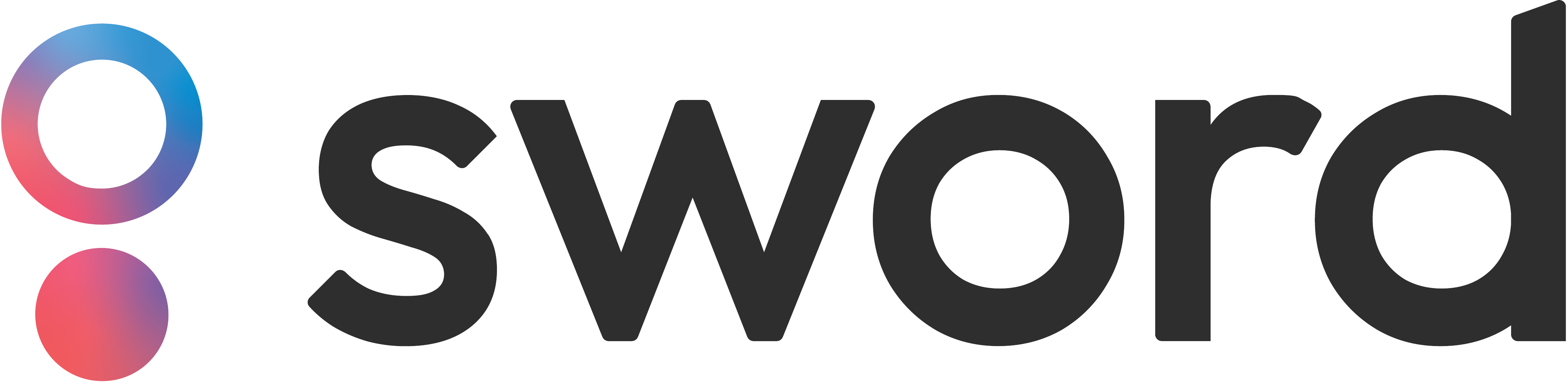}%
  \fi
}
\newcommand{\swordgradientrule}{%
  \begingroup
    \setlength{\parskip}{0pt}%
    \noindent
    \begin{tikzpicture}[baseline=0pt]
      \ifnum\swordtitlebrandvariant=0\relax
        \shade[
          shading=axis,
          left color=swordcoral,
          middle color=swordviolet,
          right color=swordblue,
          shading angle=90,
          rounded corners=0.7pt
        ] (0,0) rectangle (\linewidth,1.4pt);
      \else
        \fill[black,rounded corners=0.7pt]
          (0,0) rectangle (\linewidth,1.4pt);
      \fi
    \end{tikzpicture}\par
  \endgroup
}
\makeatletter
\renewcommand{\@maketitle}{%
  \newpage
  \null
  \vspace{-0.9cm}%
  \begingroup
    \setlength{\parindent}{0pt}%
    \raggedright
    \ifnum\swordtitleruleplacement=0\relax
      \swordgradientrule
      \vspace{0.22cm}%
    \fi
    \swordtitlelogo\par
    \ifnum\swordtitleruleplacement=1\relax
      \vspace{-0.08cm}%
      \swordgradientrule
      \vspace{0.22cm}%
    \else
      \vspace{0.25cm}%
    \fi
    {\swordtitleface\LARGE\bfseries\@title\par}
    \vspace{0.22cm}%
    {\@author\par}
  \endgroup
}
\makeatother

\begin{document}
\maketitle
\vspace{0.22cm}
{\labelfont\fontsize{10.2}{12}\selectfont Abstract\par}
\vspace{0.04cm}
{\small
Reinforcement learning for language models has moved from preference reward models to verifiable rewards, which sustain much longer optimization.
Preference models are prone to reward hacking: the policy raises their score without improving response quality, or even while degrading it.
Yet for open-ended tasks, where quality can neither be verified nor fully written down as rubrics, they remain valuable.
We introduce \methodname{}, which mitigates reward hacking online, during policy training.
Periodically and asynchronously, a language model analyzes training rollouts and writes penalties for undesirable behaviors that the reward model appears to reward; a judge applies them to every rollout.
In clinical counseling and creative writing, training on the reward model alone raises its score continuously but ultimately leads to a decline in external benchmarks; \methodname{} sustains longer training and results in improved downstream performance. We find that our method suppresses patterns incentivized by preference models that, without penalties, spread to most responses, and even to the model's reasoning, which the reward model does not explicitly score. Furthermore, we show that its gains hold in multi-turn training, and even after we retrain the reward model on the discovered behaviors.
\par}
\vspace{0.35cm}

\section{Introduction}
Reinforcement learning (RL) is emerging as the \textit{de facto} paradigm for improving large language models (LLMs), and currently takes a growing share of the compute used to train them --- in some cases matching or even exceeding pretraining~\citep{lambert2025reinforcement,karpathy2025yearinreview,deepseekai2025v32,xai2025grok4,cursor2026composer15}. This scale-up derives, in part, from a shift in reward design: earlier models were mostly trained with preference models, i.e., reward models (RMs) fitted to pairwise human preferences with the Bradley-Terry model~\citep{bradley1952rank,christiano2017preferences,ouyang2022instructgpt}, whereas verifiable rewards, such as answer checks in mathematics or unit tests in code, are now the standard~\citep{lambert2024tulu3,deepseekai2025r1}.

Verifiable rewards allow much longer optimization~\citep{deepseekai2025r1,karpathy2025yearinreview,xai2025grok4} than preference models, which are widely known to be prone to \textit{reward hacking}~\citep{skalse2022defining,wang2026rewardhacking}. As optimization progresses, the policy may discover and exploit shortcuts that do not reflect response adequacy and quality (e.g., lexical features or response length; \citealp{xia2025cognition,srivastava2025causalrubrics,wang2026rewardhacking}), thus increasing reward-model scores without improving, or even while degrading, downstream response quality.

However, not all tasks, particularly open-ended ones such as clinical counseling and creative writing, can easily be put in a form that a program can verify~\citep{jia2025writingzero,gunjal2025rubrics,huang2025rubricanchors,zhang2025vmrrlvr,wang2026rlsvr}. For these tasks, rubric-based rewards have emerged as an alternative to do away with preference models~\citep{deepseekai2025v32,deepseekai2026v4}: reward scores are derived from LLM judges running each response against a predefined set of rubrics~\citep{gunjal2025rubrics,huang2025rubricanchors,kimiteam2025k2,kimiteam2026k25,harvey2026tenet} (typically written for each task or instance; \citealp{huang2026rubicap,li2026ares}). Yet a rubric rewards only what its authors --- humans or language models --- thought or were prompted to write down~\citep{mahmoud2026rubrichacking,yang2026rubricdropout}.\footnote{Rubrics are hard to set even with clinical expertise: psychiatrists agree poorly when they rate LLM responses with a shared, calibrated rubric~\citep{jafarimeimandi2026expert}.} We argue that, for such tasks, preference models remain valuable: they can encode information that is hard to write down, and promote and penalize behaviors and properties that are hard to enumerate. And although previous work has attempted to make reward models more robust and less exploitable~\citep{srivastava2025causalrubrics,bukharin2025adversarial,coste2024ensembles,eisenstein2024helping,duan2026bayesianreward}, it is hard, if not impossible, to rely purely on reward models (and combinations thereof) during training and remain exploit-free.

As such, in this work, we focus on addressing this gap by exploring methods to mitigate reward hacking of preference models \textit{online}, adaptively during policy training. We introduce \methodname{}, a method that complements a reward model with penalties that evolve during training. Periodically, during the RL climb, a separate language model analyzes policy rollouts and their reward scores, identifies undesirable behaviors that appear to be rewarded by the RM, and writes penalties for them. A dynamic penalty buffer keeps these penalties up to date, adding new ones and discarding those that are no longer triggered by the policy at the time. Importantly, both the proposal and the buffer update run asynchronously, without interrupting policy optimization. Finally, a judge applies the buffer's penalties to each rollout, so the reward for gradient updates adapts to the reward-hacking patterns observed during training, while the RM still defines response quality.

We evaluate \methodname{} in two open-ended tasks: clinical counseling and creative writing. Our contributions are as follows:
\begin{itemize}
    \item We introduce \methodname{}, which discovers reward-hacking behaviors during policy training and penalizes them online and asynchronously, while the RM remains the primary signal of response quality (Section~\ref{sec:method}).
    \item We show that reward models are easily ``hacked''. In both tasks, RM-only training keeps raising the RM score while performance on external benchmarks declines. \methodname{} not only stops this decline: it allows longer RL climbs that lead to improved downstream performance (Section~\ref{sec:results}).
    \item We analyze how \methodname{} works in practice: which behaviors the proposer discovers, how their prevalence changes once they enter the buffer, and how widespread they are without penalties (Section~\ref{sec:penalty-analysis}).
    \item We examine two potential mitigations: multi-turn RL, where the policy holds several turns of conversation with a simulated user, and reward-model retraining, where the RM is retrained on the behaviors that \methodname{} discovered. We find that neither prevents reward hacking, and that \methodname{} reaches higher performance than RM-only training in both (Sections~\ref{sec:multi-turn} and~\ref{sec:retrained-rm}).
\end{itemize}
We will publicly release the code that led to the experiments in this paper.\footnote{We will share these resources upon acceptance.}

\section{Background}
\label{sec:background}
\begin{figure}[t]
    \centering
    \includegraphics[width=\linewidth]{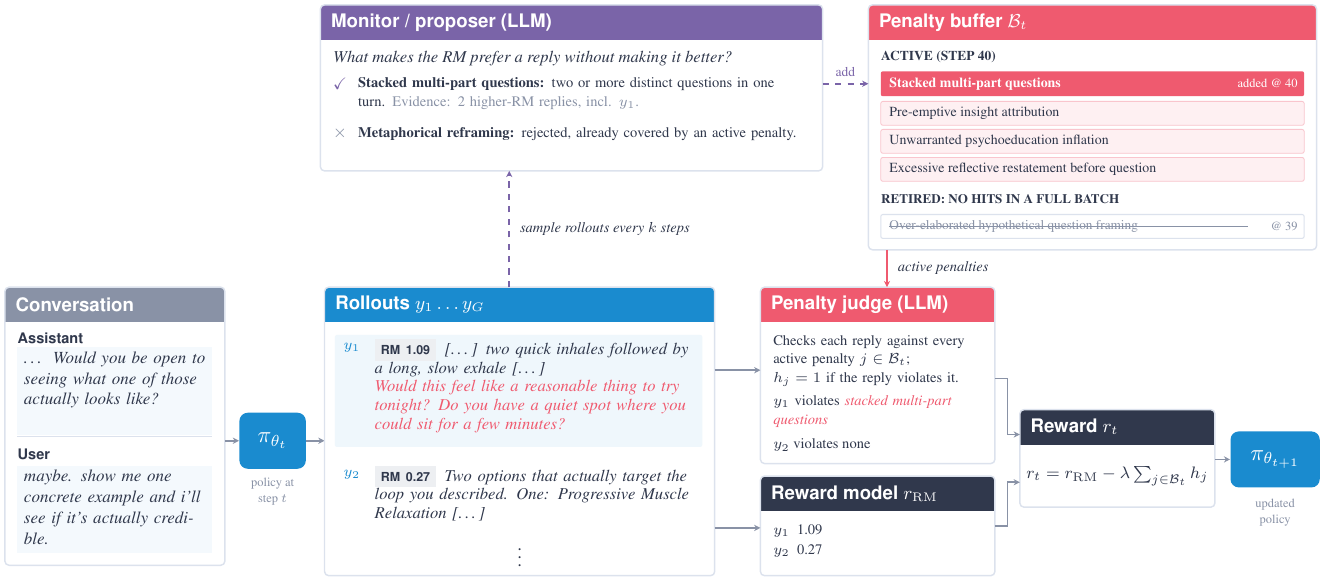}
    \caption{Overview of \methodname{}. Full details in Section~\ref{sec:method}.}
    \label{fig:method}
\end{figure}

\paragraph{Preference reward models.}
A preference reward model assigns a scalar score $r_{\mathrm{RM}}(x,y)$ to a response $y$ given an input $x$. It is trained on a dataset $\mathcal{D}$ of human preferences, where each item holds an input $x$, a preferred response $y^{+}$ and a rejected response $y^{-}$. Under the Bradley-Terry model~\citep{bradley1952rank}, the probability that $y^{+}$ is preferred over $y^{-}$ is
\begin{equation}
    P(y^{+} \succ y^{-} \mid x) = \sigma\big(r_{\mathrm{RM}}(x,y^{+}) - r_{\mathrm{RM}}(x,y^{-})\big),
    \label{eq:bradley-terry}
\end{equation}
where $\sigma$ is the logistic function. The reward model is trained to maximize the likelihood of the preferences in $\mathcal{D}$~\citep{ouyang2022instructgpt}. In this paper, we use preference models as the main reward for open-ended tasks, where a response cannot be verified automatically and such models remain attractive.

\paragraph{Preference optimization.}
Preference optimization, in particular Group Relative Policy Optimization (GRPO; \citealp{shao2024deepseekmath}) and its variants, have become a standard component on post-training pipelines of language models~\citep{deepseekai2025r1,deepseekai2025v32,liu2025drgrpo,qi2026dppo}. Under GRPO, for each input $x$, the policy $\pi_\theta$ samples a group of $G$ responses $y_1,\dots,y_G$, and each response receives a reward $r_i = r(x,y_i)$. The reward can come from a program that verifies the response~\citep{lambert2024tulu3,deepseekai2025r1}, from an LLM judge that scores it against rubrics~\citep{gunjal2025rubrics,huang2025rubricanchors}, or from a preference reward model. It compares each response with the others in its group through its \emph{advantage}, the reward of the response relative to the mean reward of the group:
\begin{equation}
    A_i = r_i - \frac{1}{G}\sum_{g=1}^{G} r_g .
    \label{eq:advantage}
\end{equation}
The policy is then updated to raise the likelihood of responses with positive advantage and to lower the likelihood of responses with negative advantage.\footnote{The original formulation also divides the advantage by the standard deviation of the group rewards; we do not. For the update, we use the default RL loss of \texttt{prime-rl} (\href{https://github.com/PrimeIntellect-ai/prime-rl/blob/cabbd132eb714b30e6be3274fe3f0766325958b9/docs/algorithms.md\#default-rl-loss}{documentation}, \href{https://github.com/PrimeIntellect-ai/prime-rl/blob/cabbd132eb714b30e6be3274fe3f0766325958b9/src/prime_rl/trainer/rl/loss.py}{code}), which combines the clipped policy-gradient term of DPPO~\citep{qi2026dppo}, a variant of GRPO, with the KL regularizer of Kimi~K2.5~\citep{kimiteam2026k25}.} In this paper, the reward $r_i$ is based on the score of a preference reward model, $r_{\mathrm{RM}}(x,y_i)$.

\section{Method}
\label{sec:method}

We illustrate our method, \methodname{}, in Figure~\ref{fig:method}. We focus exclusively on using a preference reward model (RM) as the main training signal. The method has two main components:
\begin{itemize}
\item \emph{penalty proposer}: a separate language model that identifies emerging reward-hacking patterns in training rollouts and formulates them as negative rubrics.\footnote{A penalty is not necessarily a failure mode in a single reply, but rather a behavior that may be disproportionately incentivized by the RM in a way that may render it a problem. For example, starting a reply with ``Okay'' is harmless, but starting most replies with it is not. Hence, we focus on behaviors that the proposer identifies in several rollouts.} These define behaviors to penalize in subsequent policy updates.
\item \emph{penalty buffer}: the set $\mathcal{B}_t$ of penalties that are active at training step $t$. It takes in the new penalties from the proposer and drops those that the policy no longer triggers.
\end{itemize}
Below, we first define the training reward, and then describe each component.

Every $k$ training steps, the proposer analyzes a sample of rollouts from the previous step and proposes new penalties, which are added to the penalty buffer $\mathcal{B}_t$. For each response $y$ to an input $x$ (the context, prompt or conversation history), an LLM judge checks whether any active penalty applies. The resulting training reward is
\begin{equation}
    r_t(x,y) = r_{\mathrm{RM}}(x,y)
    - \lambda \sum_{j \in \mathcal{B}_t} h_j(x,y),
    \label{eq:penalized-reward}
\end{equation}
where $h_j(x,y) \in \{0,1\}$ is the judge's verdict on whether response $y$ exhibits the behavior of penalty $j$, and $\lambda > 0$ is the penalty weight. This reward replaces $r(x,y_i)$ in Equation~\ref{eq:advantage}.

\paragraph{Penalty proposer.}
Prior work shows that LLMs can monitor model behavior to detect reward hacking and other forms of misbehavior\citep{baker2025monitoring,korbak2025monitorability}. Here, we use a language model to discover reward-hacking patterns in training rollouts and convert them into explicit criteria for detecting these patterns in future rollouts. Specifically, at each proposal step, the proposer receives 32 rollouts randomly sampled from the batch generated at the previous training step, their RM scores, and the active penalties $\mathcal{B}_t$. The rollouts are labeled according to whether their advantage is positive or negative (Equation~\ref{eq:advantage}, computed with the RM score). This helps the proposer identify which responses the optimization reinforces and focus on reward-hacking patterns in positive-advantage responses. We do this because identifying patterns in negative-advantage responses is less useful, since these responses are already discouraged by the RM.

Each candidate penalty is expressed as a negative rubric with a short title, a description of the pattern, and references to sampled responses that illustrate it.\footnote{Before a candidate enters the buffer, we run a second LLM pass that compares it against the active penalties, and we discard it if it describes a behavior that an active penalty already covers. This prevents duplicate penalties from penalizing the same behavior multiple times in Equation~\ref{eq:penalized-reward}.} Appendix~\ref{app:penalty-prompts} provides the prompts.

\paragraph{Penalty buffer.}
The buffer stores the active penalties used by the judge at each training step. A penalty is removed only if it is not triggered by any of the sampled rollouts in a given step. Each proposal step adds at most one new penalty.
 During training, a judge checks each rollout for the behaviors described by the active penalties. For efficiency purposes, particularly as $|\mathcal{B}_t|$ grows, all penalties are evaluated in a single judge pass.

\paragraph{Asynchronous updates.}
We run penalty proposal and buffer management asynchronously to avoid interrupting policy optimization. Training uses the current set of active penalties while new penalties are proposed in the background. Once a buffer update is complete, the updated set takes effect at the next batch of rollouts. This allows us to adapt penalties during training without waiting for proposal or buffer management to finish.

\section{Experimental Setup}
\label{sec:experimental-setup}
We evaluate \methodname{} on two open-ended tasks: clinical counseling and creative writing. We match each task with a relevant reward model, which provides the training signal, and with a relevant external benchmark, on which we evaluate policy checkpoints. We use the benchmark as an estimate of the true response quality. This allows us to tell reward hacking apart from downstream quality improvements: as in prior work on learned rewards~\citep{gao2022overoptimization,yang2026rubricdropout}, we take a benchmark score that declines while the RM score keeps increasing as the signal of reward hacking. For each task, we compare RM-only training, where the reward is the RM score $r_{\mathrm{RM}}(x,y)$, with \methodname{} (Equation~\ref{eq:penalized-reward}).

\paragraph{Clinical counseling.}
In this task, the policy acts as a clinical counselor: given a multi-turn conversation with a patient, the model is tasked with writing the next reply. The prompts for reinforcement learning derive from a private dataset of 10,000 conversation samples. The reward model is a proprietary Bradley--Terry model (Section~\ref{sec:background}) trained on preferences derived from annotations from clinical psychologists: a higher score indicates a reply that a psychologist is more likely to prefer. We evaluate on MindEval~\citep{pombal2025mindeval}, a benchmark of multi-turn mental health support conversations. Appendix~\ref{app:evaluation-setup} describes the exact setup.

\paragraph{Creative writing.}
In this task, the policy writes a short story for a given prompt. We use the prompts and the reward model of LitBench~\citep{fein2025litbench}. We train on the 12,414 unique prompts of \href{https://huggingface.co/datasets/SAA-Lab/LitBench-Train}{\texttt{LitBench-Train}}. The reward model is \href{https://huggingface.co/SAA-Lab/Llama8B-CreativeWritingVerifier}{\texttt{Llama8B-CreativeWritingVerifier}}, a Bradley--Terry model trained on human preferences over stories. We evaluate policy checkpoints on EQ-Bench Creative Writing v3~\citep{paech2023eqbench}; Appendix~\ref{app:evaluation-setup} describes the exact setup.

\begin{figure}[t]
    \centering
    \begin{subfigure}[t]{0.49\textwidth}
        \centering
        \includegraphics[width=\linewidth]{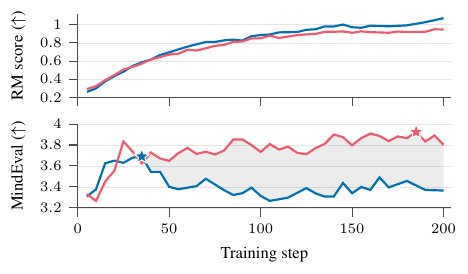}
        \caption{Clinical counseling}
        \label{fig:training-curves-clinical}
    \end{subfigure}\hfill
    \begin{subfigure}[t]{0.49\textwidth}
        \centering
        \includegraphics[width=\linewidth]{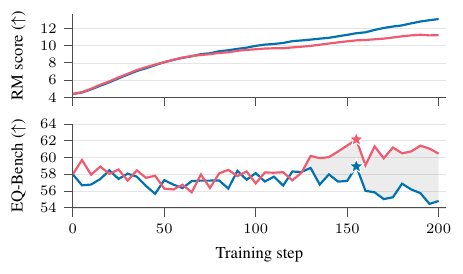}
        \caption{Creative writing}
        \label{fig:training-curves-writing}
    \end{subfigure}

    \smallskip
    \includegraphics{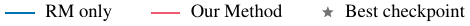}
    \caption{RM score and benchmark score over training, for RM-only training and our method.}
    \label{fig:training-curves}
\end{figure}
\begin{table}[t]
\centering
\small
\setlength{\tabcolsep}{6pt}
\begin{tabular}{l lllr lllr}
\toprule
 & \multicolumn{4}{c}{\textbf{Clinical counseling}} & \multicolumn{4}{c}{\textbf{Creative writing}} \\
 & \multicolumn{4}{c}{MindEval} & \multicolumn{4}{c}{EQ-Bench CW v3} \\
\cmidrule(lr){2-9}
\textbf{Run} & Peak & End & Final & RM score & \hspace{14pt}Peak & End & Final & RM score \\
\midrule
RM only & 3.69 & 3.39 & 3.37 & 1.00 & \hspace{14pt}58.9 & 56.0 & 54.8 & 12.19 \\
\rowcolor{OursRow}\methodname{} & \textbf{3.92}\gain{+0.23} & \textbf{3.84}\gain{+0.45} & \textbf{3.80}\gain{+0.43} & 0.92 & \hspace{14pt}\textbf{62.1}\gain{+3.2} & \textbf{60.8}\gain{+4.8} & \textbf{60.5}\gain{+5.7} & 10.93 \\
\bottomrule
\end{tabular}
\caption{Main results. \emph{Peak}: best checkpoint. \emph{End}: mean over the checkpoints evaluated for both runs in steps 150--200. \emph{Final}: step 200. \emph{RM score}: mean over the \emph{End} checkpoints. Subscripts: difference from RM only.}
\label{tab:headline-results}
\end{table}

\paragraph{Training configuration.}
For each task, we train a \href{https://huggingface.co/Qwen/Qwen3.5-35B-A3B}{Qwen3.5-35B-A3B}~\citep{qwenteam2026qwen35}\footnote{This is a released model that has already undergone post-training. For clinical counseling, it has gone further supervised fine-tuning for clinical appropriateness.} model for 200 steps with the RL setup described in Section~\ref{sec:background}. Each step uses a batch of 512 rollouts: 32 prompts with a group of $G = 16$ responses each, sampled at temperature 1. We update all the weights of the policy with AdamW, with a constant learning rate of $10^{-6}$ and a weight decay of 0.01. Rollout generation and training run asynchronously, and we discard rollouts that are more than eight steps behind the current policy. 

\paragraph{Penalty configuration.}
The penalty proposer is \href{https://huggingface.co/zai-org/GLM-5.2}{GLM~5.2}~\citep{glm5team2026glm5}. Because the proposer runs asynchronously and does not slow down training, we can use a large, capable model to discover reward-hacking patterns. It runs every $k = 5$ training steps on 32 rollouts from the previous batch: two rollouts sampled at random from each of 16 groups. A candidate penalty is kept only if the proposer finds evidence for it in at least two groups. The judge that returns the verdicts $h_j$ is \href{https://huggingface.co/Qwen/Qwen3.6-35B-A3B}{Qwen3.6-35B-A3B}~\citep{qwenteam2026qwen36}. In contrast to the proposer, the judge runs on every rollout, so we use a smaller model that keeps judging fast and avoids slowing down training. We set the penalty weight to $\lambda = 0.5$ in Equation~\ref{eq:penalized-reward}.

\section{Results and Analysis}\label{sec:results}

\subsection{\methodname{} mitigates reward hacking}

We can see that, for both clinical counseling and creative writing, RM-only training leads to steady increases in the respective RM scores during training, usually dominating those of our method (Figure~\ref{fig:training-curves}). However, importantly, response quality (as measured by the performance on MindEval and EQ-Bench, respectively) starts degrading --- this is the signal of reward hacking. This is not the case with our method, where response quality keeps improving for much longer. \methodname{} not only allows us to extend training: it keeps optimizing for the reward model without sacrificing performance. In fact, in both tasks, it ends with a benchmark score superior to that of RM-only training, not only at its end --- where hacking is most prominent --- but also at its peak (Table~\ref{tab:headline-results}). One hypothesis is that the penalties keep the policy from being steered towards artifacts or spurious features that the RM ``likes''. We will analyze this next.

\begin{figure}[t]
    \centering
    \includegraphics[width=\linewidth]{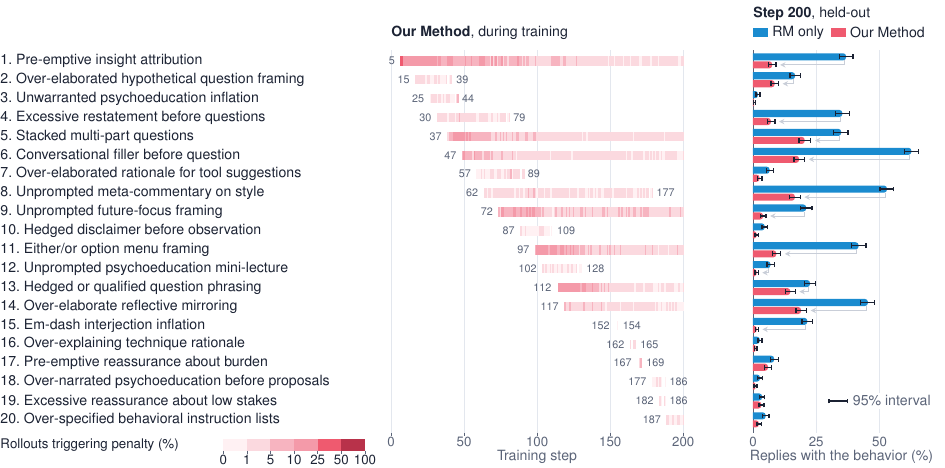}
    \caption{The 20 penalties discovered in clinical counseling: their prevalence in training rollouts while in the buffer, with the steps at which each was added and retired (left), and in held-out replies at the end of training (right).}
    \label{fig:clinical-penalty-timeline}
\end{figure}

\subsection{What does the proposer discover?}
\label{sec:penalty-analysis}

Now we turn to analyzing the behaviors that were being penalized by our method during training.

Figure~\ref{fig:clinical-penalty-timeline} illustrates the penalty buffer during training for the clinical counseling task.\footnote{Table~\ref{tab:clinical-penalty-descriptions} provides the full description of each penalty. For creative writing, Figure~\ref{fig:writing-penalty-timeline} in App.~\ref{app:penalty-lifecycles} shows the same figure, and App.~\ref{app:penalty-descriptions} provides the descriptions.} 

Some behaviors are common when first discovered but become substantially less frequent after being penalized. The most common ones, such as those targeted by penalties 1, 5, and 6, do not disappear completely: they persist at low rates until the end of training, and their penalties stay in the buffer. In contrast, less frequent behaviors, such as those targeted by penalties 15--19, disappear within a few training steps, and their penalties are retired. Because some behaviors persist, the buffer tends to grow during training.

The penalties range from surface-level, lexical patterns to content adequacy and quality.

\paragraph{The same response patterns are common and sticky under RM-only training.}
To assess whether these patterns also emerge without penalties, we evaluate the outputs of the RM-only policy at the end of training against all criteria discovered by our method, with the same judge and the same held-out prompts as for \methodname{}. For example, in clinical counseling (right panel of Figure~\ref{fig:clinical-penalty-timeline}), over 90\% of RM-only replies trigger at least one criterion, with over four criteria per reply on average, against fewer than two for \methodname{}. These results suggest that the policy heavily exploits these patterns when trained without penalties. They also suggest that the patterns are sticky: under RM-only training nothing penalizes them, and in clinical counseling the average number of criteria per reply almost triples from the start to the end of training.

\begin{table}[t]
\centering
\small
\setlength{\tabcolsep}{4pt}
\begin{tabular}{l cccccc}
\toprule
 & Clinical accuracy & Ethical \& prof. & Assessment & Therapeutic rel. & AI-specific &  \\
\textbf{Run} & \& competence & conduct & \& response & \& alliance & comm. quality & Overall \\
\midrule
RM only, end & 3.54 & 3.98 & 3.37 & 3.51 & 2.52 & 3.39 \\[3pt]
\hdashline\noalign{\vskip 3pt}
RM only, peak & 3.66\gainlight{+0.12} & 4.13\gainlight{+0.15} & 3.60\gainlight{+0.23} & 3.76\gainlight{+0.25} & 3.29\gainlight{+0.77} & 3.69\gainlight{+0.30} \\
\rowcolor{OursRow}\methodname{}, end & \textbf{3.73}\gain{+0.19} & \textbf{4.32}\gain{+0.34} & \textbf{3.72}\gain{+0.35} & \textbf{3.79}\gain{+0.28} & \textbf{3.66}\gain{+1.14} & \textbf{3.84}\gain{+0.45} \\
\bottomrule
\end{tabular}
\caption{MindEval scores by axis. \emph{Peak} and \emph{End} as in Table~\ref{tab:headline-results}. Subscripts: difference from RM only at the end.}
\label{tab:mindeval-axes}
\end{table}

\begin{figure}[t]
    \centering
\begingroup
\definecolor{patternRM}{RGB}{26,139,207}
\definecolor{patternOurs}{RGB}{239,90,111}
\definecolor{patternMuted}{RGB}{91,100,120}
\begin{tikzpicture}
\begin{groupplot}[
    group style={group size=3 by 1,horizontal sep=31pt},
    width=0.255\linewidth,height=78pt,scale only axis,
    xmin=20.40973,xmax=190.49059,
    xtick={20.40973,62.92993,105.45016,147.97037,190.49059},
    xticklabels={0,50,100,150,200},
    xlabel={Training step},
    tick label style={font=\sffamily\fontsize{6.5}{7.5}\selectfont,text=patternMuted},
    label style={font=\sffamily\fontsize{7}{8}\selectfont,text=patternMuted},
    title style={font=\sffamily\fontsize{8}{9.5}\selectfont,align=center,yshift=2pt},
    axis x line*=bottom,axis y line*=left,
    axis line style={draw=patternMuted!65,line width=0.35pt},
    y axis line style={draw=none},ytick style={draw=none},
    tick style={draw=patternMuted!65,line width=0.35pt},
    major tick length=2pt,
    ymajorgrids=true,grid style={draw=patternMuted!15,line width=0.35pt},
    clip=false,
    every axis plot/.append style={line width=1.0pt,no marks,line join=round,line cap=round},
    xlabel style={at={(axis description cs:0.5,-0.21)},anchor=north},
]
\nextgroupplot[title={Clinical counseling\\Replies starting with ``Okay''},ymin=-45.35506,ymax=0,ytick={-45.35506,-30.24606,-15.10896,0},yticklabels={0\%,20\%,40\%,60\%}]
\addplot[patternRM] coordinates {
(24.66158,-44.36282)
(28.91385,-43.90909)
(33.16571,-43.76764)
(37.41797,-42.94539)
(41.66983,-43.76764)
(45.92168,-42.94539)
(50.17395,-42.40688)
(54.42581,-41.64128)
(58.6781,-40.96133)
(62.92993,-40.13908)
(67.1818,-39.09018)
(71.43407,-37.10612)
(75.68591,-36.28386)
(79.9382,-34.10129)
(84.19003,-32.06013)
(88.4419,-32.74051)
(92.69418,-32.51385)
(96.94601,-30.84123)
(101.1983,-29.48047)
(105.45016,-24.34972)
(109.702,-23.44267)
(113.95428,-24.8601)
(118.20613,-21.0049)
(122.4584,-19.41748)
(126.71027,-16.10077)
(130.9621,-12.47266)
(135.21439,-9.75111)
(139.46625,-6.49152)
(143.71852,-7.39856)
(147.97037,-8.24892)
(152.22221,-7.56854)
(156.4745,-6.43486)
(164.97862,-6.88815)
(173.48233,-4.84743)
(181.98645,-3.03294)
(190.49059,-0.31184)
};
\node[anchor=south east,xshift=-1pt,yshift=1pt,inner sep=1pt,fill=white,font=\sffamily\fontsize{6.5}{7.5}\selectfont] at (axis cs:190.49059,-0.31184) {60\%};
\addplot[patternOurs] coordinates {
(24.66158,-44.30614)
(28.91385,-44.13615)
(33.16571,-43.62576)
(37.41797,-43.54056)
(41.66983,-43.3139)
(45.92168,-43.08727)
(50.17395,-41.64128)
(54.42581,-40.67758)
(58.6781,-40.28094)
(62.92993,-39.68535)
(67.1818,-40.73425)
(71.43407,-42.265)
(75.68591,-42.32166)
(79.9382,-43.08727)
(84.19003,-42.77539)
(88.4419,-43.45578)
(92.69418,-43.08727)
(96.94601,-42.86018)
(101.1983,-42.32166)
(105.45016,-42.09502)
(109.702,-42.03836)
(113.95428,-43.08727)
(118.20613,-42.49165)
(122.4584,-43.17203)
(126.71027,-42.49165)
(130.9621,-42.86018)
(135.21439,-43.3139)
(139.46625,-43.2287)
(143.71852,-43.3139)
(147.97037,-43.45578)
(152.22221,-43.85242)
(156.4745,-43.17203)
(160.72635,-43.45578)
(164.97862,-43.17203)
(169.23047,-43.68243)
(173.48233,-43.76764)
(177.7346,-43.45578)
(181.98645,-43.08727)
(186.23872,-43.62576)
(190.49059,-43.08727)
};
\node[anchor=south east,xshift=-1pt,yshift=1pt,inner sep=1pt,fill=white,font=\sffamily\fontsize{6.5}{7.5}\selectfont] at (axis cs:190.49059,-43.08727) {3\%};
\draw[patternMuted!75,densely dashed,line width=0.5pt]
    (axis cs:60.37883,-45.35506) -- (axis cs:60.37883,0);
\node[anchor=north west,align=left,inner sep=2pt,
    font=\sffamily\fontsize{6}{7}\selectfont,text=patternMuted]
    at (axis cs:60.37883,0) {Penalty added\\(step 47)};
\nextgroupplot[title={Creative writing\\Stories containing ``galactic''},ymin=-112.82018,ymax=-67.46553,ytick={-112.82018,-90.14264,-67.46553},yticklabels={0\%,50\%,100\%}]
\addplot[patternRM] coordinates {
(24.66158,-111.74315)
(28.91385,-111.62982)
(33.16571,-111.62982)
(37.41797,-111.74315)
(41.66983,-111.5165)
(45.92168,-111.5165)
(50.17395,-111.17609)
(54.42581,-111.37463)
(58.6781,-111.28941)
(62.92993,-111.14798)
(67.1818,-111.14798)
(71.43407,-111.14798)
(75.68591,-111.06276)
(79.9382,-111.0061)
(84.19003,-110.8361)
(88.4419,-110.7228)
(92.69418,-111.0061)
(96.94601,-110.60904)
(101.1983,-110.69424)
(105.45016,-110.32571)
(109.702,-110.26906)
(113.95428,-110.7228)
(118.20613,-110.18384)
(122.4584,-109.87242)
(126.71027,-109.19203)
(130.9621,-108.59644)
(135.21439,-108.56831)
(139.46625,-107.46275)
(143.71852,-104.82643)
(147.97037,-100.88644)
(152.22221,-98.25012)
(156.4745,-94.96198)
(160.72635,-85.60747)
(164.97862,-76.22443)
(169.23047,-70.7818)
(173.48233,-69.22293)
(177.7346,-68.51443)
(181.98645,-68.59921)
(186.23872,-67.63553)
(190.49059,-67.74884)
};
\node[anchor=south east,xshift=-1pt,yshift=1pt,inner sep=1pt,fill=white,font=\sffamily\fontsize{6.5}{7.5}\selectfont] at (axis cs:190.49059,-67.74884) {99\%};
\addplot[patternOurs] coordinates {
(24.66158,-111.68648)
(28.91385,-111.31796)
(33.16571,-111.60127)
(37.41797,-111.14798)
(41.66983,-111.40274)
(45.92168,-111.23274)
(50.17395,-110.7228)
(54.42581,-110.77945)
(58.6781,-110.9209)
(62.92993,-111.23274)
(67.1818,-110.9209)
(71.43407,-111.09131)
(75.68591,-111.0061)
(79.9382,-111.06276)
(84.19003,-110.77945)
(88.4419,-111.17609)
(92.69418,-111.09131)
(96.94601,-110.7228)
(101.1983,-110.86424)
(105.45016,-111.0061)
(109.702,-111.09131)
(113.95428,-111.0061)
(118.20613,-111.0061)
(122.4584,-111.17609)
(126.71027,-111.40274)
(130.9621,-111.06276)
(135.21439,-111.09131)
(139.46625,-111.45984)
(143.71852,-111.06276)
(147.97037,-111.06276)
(152.22221,-111.31796)
(156.4745,-110.9209)
(160.72635,-111.31796)
(164.97862,-111.28941)
(169.23047,-110.7228)
(173.48233,-111.14798)
(177.7346,-111.28941)
(181.98645,-110.8361)
(186.23872,-110.86424)
(190.49059,-110.8361)
};
\node[anchor=south east,xshift=-1pt,yshift=1pt,inner sep=1pt,fill=white,font=\sffamily\fontsize{6.5}{7.5}\selectfont] at (axis cs:190.49059,-110.8361) {4\%};
\nextgroupplot[title={Creative writing (reasoning)\\Mentions of r/WritingPrompts},ymin=-180.2857,ymax=-134.93063,ytick={-180.2857,-157.60817,-134.93063},yticklabels={0\%,50\%,100\%}]
\addplot[patternRM] coordinates {
(24.66158,-179.77531)
(28.91385,-179.97386)
(33.16571,-179.83199)
(37.41797,-179.83199)
(41.66983,-179.09494)
(45.92168,-178.52832)
(50.17395,-178.6983)
(54.42581,-178.47165)
(58.6781,-178.415)
(62.92993,-178.415)
(67.1818,-178.24457)
(71.43407,-179.09494)
(75.68591,-178.6983)
(79.9382,-177.79126)
(84.19003,-175.52347)
(88.4419,-176.26051)
(92.69418,-177.11089)
(96.94601,-174.55978)
(101.1983,-173.39754)
(105.45016,-172.4053)
(109.702,-171.83824)
(113.95428,-171.21452)
(118.20613,-166.08377)
(122.4584,-166.67937)
(126.71027,-160.47115)
(130.9621,-164.24158)
(135.21439,-158.79897)
(139.46625,-155.85078)
(143.71852,-149.8411)
(147.97037,-151.88225)
(152.22221,-152.30743)
(156.4745,-152.22221)
(160.72635,-152.93115)
(164.97862,-159.7341)
(169.23047,-159.87598)
(173.48233,-166.05565)
(177.7346,-164.58156)
(181.98645,-160.55635)
(186.23872,-159.7341)
(190.49059,-153.75298)
};
\node[anchor=south east,xshift=-1pt,yshift=1pt,inner sep=1pt,fill=white,font=\sffamily\fontsize{6.5}{7.5}\selectfont] at (axis cs:190.49059,-153.75298) {59\%};
\addplot[patternOurs] coordinates {
(24.66158,-179.97386)
(28.91385,-179.83199)
(33.16571,-179.7472)
(37.41797,-179.60533)
(41.66983,-179.29346)
(45.92168,-177.7346)
(50.17395,-175.06973)
(54.42581,-174.44644)
(58.6781,-173.93605)
(62.92993,-173.48233)
(67.1818,-174.61644)
(71.43407,-176.28862)
(75.68591,-177.79126)
(79.9382,-179.15202)
(84.19003,-179.88864)
(88.4419,-180.05907)
(92.69418,-179.88864)
(96.94601,-179.7472)
(101.1983,-179.88864)
(105.45016,-179.7472)
(109.702,-179.88864)
(113.95428,-180.11572)
(118.20613,-179.88864)
(122.4584,-179.7472)
(126.71027,-180.00241)
(130.9621,-179.91719)
(135.21439,-179.83199)
(139.46625,-179.09494)
(143.71852,-178.24457)
(147.97037,-178.01791)
(152.22221,-179.60533)
(156.4745,-179.43535)
(160.72635,-179.83199)
(164.97862,-179.83199)
(169.23047,-179.7472)
(173.48233,-179.88864)
(177.7346,-180.05907)
(181.98645,-179.7472)
(186.23872,-179.52013)
(190.49059,-178.92496)
};
\node[anchor=south east,xshift=-1pt,yshift=1pt,inner sep=1pt,fill=white,font=\sffamily\fontsize{6.5}{7.5}\selectfont] at (axis cs:190.49059,-178.92496) {3\%};
\end{groupplot}
\node[anchor=north,inner sep=0pt,font=\sffamily\fontsize{7.5}{9}\selectfont]
    at ($(group c1r1.south)!0.5!(group c3r1.south)+(0,-34pt)$) {%
    \tikz[baseline=-0.5ex]\draw[patternRM,line width=1pt] (0,0)--(12pt,0);\enspace RM only
    \hspace{14pt}%
    \tikz[baseline=-0.5ex]\draw[patternOurs,line width=1pt] (0,0)--(12pt,0);\enspace Our Method};
\end{tikzpicture}
\endgroup
    \caption{Repeated patterns over training, on held-out prompts, for RM-only training and our method.}
    \label{fig:pattern-trajectories}
\end{figure}
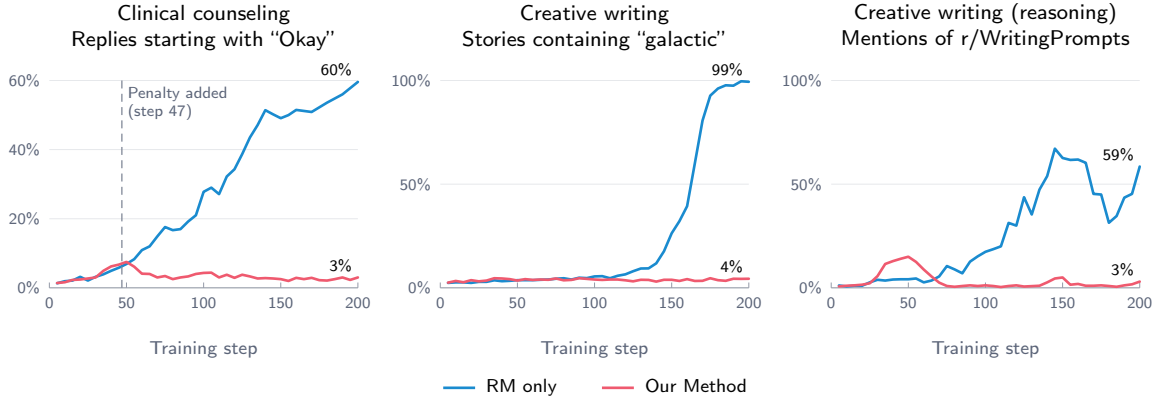

\paragraph{Without penalties, one pattern can spread to most responses.}
One of the main problems with reward hacking is that patterns start to show up in a large number of responses, even under different contexts. As we have noted in Section~\ref{sec:method}, one behavior in itself (e.g., a stacked multi-part question in clinical counseling) is not necessarily damaging to the good conduct of a session; it becomes a problem, however, when the same pattern shows up in lots of different contexts.\footnote{Policies are typically trained on one user-facing reply at a time~\citep{shani2024multiturn,wu2025collabllm}. This motivates us to explore multi-turn RL in Section~\ref{sec:multi-turn}, where the reward model also sees the earlier replies of the policy.} Repeated patterns across the outputs of a model are well documented~\citep{kirk2024rlhf,xu2025echoes,paech2026antislop}. Identifying a persistent behavior can be more or less difficult, with some being more evident and others requiring many different scenarios to surface~\citep{gupta2025bloom}. We analyze this by running different training checkpoints on a set of held-out prompts.\footnote{In clinical counseling, 1,000 conversations outside the training data; in creative writing, the 971 prompts of \texttt{LitBench-Test}, none of them in \texttt{LitBench-Train}.} For example, we found that, in clinical counseling, the share of replies that start with ``Okay'' grows from almost none to about 60\% under RM-only training. Under \methodname{}, it grows in the same way until a penalty that covers it \textit{explicitly} enters the buffer and starts penalizing these replies. In creative writing, the RM-only policy collapses to one theme regardless of the prompt: at the end of training, almost all of its stories contain the word ``galactic''~(see Figure~\ref{fig:pattern-trajectories}).

\paragraph{Hacking patterns are \textit{also} present in the model reasoning.}
Reward hacking is typically analyzed in final responses only, as most reward models score the final response and ignore the reasoning. However, in autoregressive generation, the reasoning is part of the context that conditions the final response, so patterns that are rewarded in responses may transfer to the reasoning. This matters because reasoning traces are used to monitor and interpret models~\citep{baker2025monitoring,korbak2025monitorability}, and collapsed reasoning may hurt both monitoring and generalization. We found that repeated patterns indeed emerge not only in the final response but also in the reasoning traces. For example, under RM-only training in creative writing, the share of reasoning traces that name the source of the prompts (r/WritingPrompts) grows from 1\% to 59\% at the end of training (see Figure~\ref{fig:pattern-trajectories}).

\paragraph{Responses can combine multiple patterns.}
RM-only responses are also longer and tend to combine several of these patterns, such as starting with ``Okay'' and ending with a question that offers two options (Figures~\ref{fig:clinical-response-example-1}--\ref{fig:clinical-response-example-4} compare rollouts from both policies). MindEval captures these patterns in one of its axes (Table~\ref{tab:mindeval-axes}). Under RM-only training, the largest decline is in AI-Specific Communication Quality, the axis that rates whether replies are natural and free of repeated phrases and of templated or formulaic structure. \methodname{} is ahead on every axis, and by far the most on this one.

\Needspace{0.34\textheight}
\subsection{Multi-Turn RL}
\label{sec:multi-turn}
\begin{wrapfigure}{r}{0.42\textwidth}
    \vspace{-8pt}
    \centering
    \includegraphics[width=\linewidth]{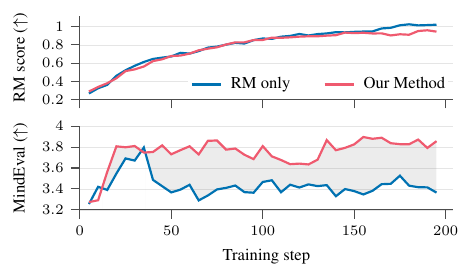}
    \captionsetup{font=footnotesize,skip=3pt}
    \caption{RM score and MindEval in multi-turn RL.}
    \label{fig:multi-turn-training-curves}
    \vspace{-8pt}
\end{wrapfigure}
So far, we have trained with single-turn RL: the policy writes one response to a conversation history taken from the dataset, and the reward model scores that response. Policies trained on such next-turn rewards are known to optimize the immediate turn at the expense of the conversation as a whole~\citep{shani2024multiturn,wu2025collabllm}. Multi-turn RL changes two things. First, the policy must continue its own conversations (and potentially ``hacked'' turns), so the reward model sees histories that contain the policy's own responses. For example, we observed that replies frequently began with ``Okay''; if this repetition appears in the conversation history, the reward model may be able to penalize it. Second, the policy is optimized on a signal that is spread over a whole trajectory, and is therefore much more dispersed than in single-turn training. We ask whether reward hacking still occurs under these conditions.

\paragraph{Setup.}
We focus exclusively on clinical counseling, for its conversational nature. Each rollout starts from a conversation in the dataset. The policy writes a reply, a user simulator (DeepSeek 4.1 Flash) writes the next message of the patient, and the policy replies again, for up to four policy replies per rollout. The reward model scores each policy reply given the conversation up to that reply, and the reward of the rollout is the mean of these scores. We compute one advantage per rollout, relative to its group, and apply it to the tokens of all policy replies in the rollout. With \methodname{}, the judge checks every policy reply against the active penalties, and the rollout is penalized once for each distinct penalty that any of its replies triggers. The remaining setup is similar to that of the original clinical counseling described in Section~\ref{sec:experimental-setup}.

\paragraph{Results.}
Figure~\ref{fig:multi-turn-training-curves} shows the training and evaluation curves. The curves are similar to those observed in single-turn training: under RM-only training, MindEval performance initially improves but then declines while RM scores continue to increase. With our method, MindEval scores remain high during training. These results suggest that our method is also effective in multi-turn environments under user simulation and that multi-turn RL alone does not prevent reward hacking.

\paragraph{Similar penalties emerge in single-turn and multi-turn training.}

We found that the penalties discovered during the multi-turn run (see Appendix~\ref{app:multi-turn-penalties}) are partly the same as those of single-turn training, including stacked questions and option-based question framing. Even in a multi-turn setting, the reward model remains easy to game, and the policy exploits the same patterns observed in single-turn training.

\Needspace{0.34\textheight}
\subsection{Can additional training data correct reward model biases?}
\label{sec:retrained-rm}
\begin{wrapfigure}{r}{0.42\textwidth}
    \vspace{-8pt}
    \centering
    \includegraphics[width=\linewidth]{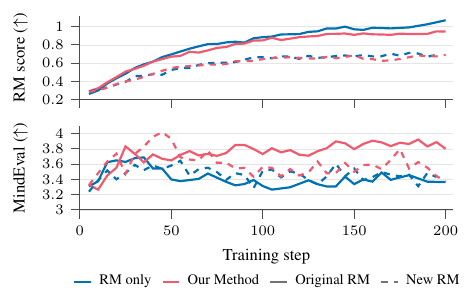}
    \captionsetup{font=footnotesize,skip=3pt}
    \caption{RM score and MindEval with the new RM.}
    \label{fig:reward-model-comparison}
    \vspace{-8pt}
\end{wrapfigure}
One hypothesis for why reward models are hacked is that they are not robust enough: they reward spurious features, and better training data or better modeling would remove these biases~\citep{srivastava2025causalrubrics,bukharin2025adversarial,coste2024ensembles,eisenstein2024helping,duan2026bayesianreward}. \methodname{} produces what such a fix needs: a list of the behaviors that the policy exploited, and rollouts that show them. Here, we investigate whether we can exploit these penalties and rollouts to train a new reward model that can potentially be more robust to the exploits found during training.

\paragraph{Setup.}

Consistent with previous analysis, we conduct this experiment only in the clinical environment. To construct training data targeting the behaviors listed in Table~\ref{tab:clinical-penalty-descriptions}, we use rollouts generated during training (we have at least eight rollouts per prompt). Then, we select as the preferred response one that triggers no penalties and has RM scores above the 75th percentile, ensuring that it is a good response according to the original RM. We then pair each preferred response with a response to the same prompt that triggers at least one penalty. This yields approximately 15,000 new pairs, which we add to the original RM training set of approximately 30,000 samples. The original and new reward models achieve similar mean accuracy across the test set.\footnote{Mean accuracy on the reward-model test set: 78.48\% for the original RM and 78.61\% for the new RM.}

\paragraph{The retrained reward model remains vulnerable to reward hacking.}
Figure~\ref{fig:reward-model-comparison} shows the training curves and MindEval scores obtained with the newly trained RM. The curves with new RM exhibits the same pattern as before: MindEval scores initially increase and then decline, even as RM scores continue to rise; we have reward hacking. Table~\ref{tab:new-rm-penalties} lists the penalties discovered during this run, which are mostly different from those found with the original reward model.

These results support our motivation: even after correcting specific biases in the reward model, the policy is likely to find other ways to exploit it so we need a method that discover and penalize this problems during training. Our approach is also less costly, as it does not require retraining the RM or restarting policy training.

\section{Conclusions}
We introduced \methodname{}, which complements a preference reward model with penalties that are discovered and updated during policy training. In clinical counseling and creative writing, it mitigates reward hacking: it sustains longer training and reaches higher downstream response quality than RM-only training. Our analysis shows that our method suppresses patterns incentivized by the preference model that, without penalties, spread to most responses, and even to the model's reasoning (which is not explicitly evaluated by the preference model). We also find that responses under our method are more natural and less formulaic than under RM-only training. We further show that neither multi-turn training nor retraining the reward model on the discovered behaviors prevents reward hacking, and that, under both contexts, our method reaches higher downstream performance than RM-only training.

\section*{Limitations}

Our work has some limitations that should be considered. First, we study reward hacking when training in a relatively simple setup: a single task with environments that are similar across all training data. An interesting direction for future work is to extend our approach to multi-environment training, where interactions across environments may introduce more complex reward-hacking behaviors that might be more challenging to detect and mitigate.

Furthermore, our method can only detect what models explicitly generate across a series of rollouts. As such, it cannot preemptively prevent hacking from occurring. It can only do so when it is already showing in model generations. Finding traces of hacking that are \textit{intrinsic} to the model, i.e., present in its internal representations before they show in its generations, is also an interesting direction for future work.

Finally, although penalty proposal and buffer updates are performed asynchronously, applying the penalty judge to every rollout adds computational overhead, although we attempt to minimize this by using a small model as the judge.

\section*{Ethical Considerations}
Clinical counseling is one of the tasks studied in this work, where the model acts as a clinical counselor in a multi-turn conversation with a patient. The training data for this task consist of a private dataset of real human conversations , which raises important ethical considerations about privacy and confidentiality. To protect patient privacy, we anonymized the data and will not make it publicly available.



%
{\small
\bibliographystyle{plainnat}
\bibliography{custom}
}

\clearpage
\appendix
\crefalias{section}{appendix}
\crefalias{subsection}{appendix}
\crefalias{subsubsection}{appendix}
\setcounter{secnumdepth}{-2}
\part{Appendix}
\setcounter{secnumdepth}{3}
\etocsetnexttocdepth{subsection}
\localtableofcontents
\clearpage

\section{Evaluation Setup}
\label{app:evaluation-setup}

\paragraph{MindEval.}
MindEval~\citep{pombal2025mindeval} evaluates a model in simulated multi-turn conversations. The model under evaluation acts as the counselor, a second language model simulates the patient from a profile, and a judge model rates each conversation from 1 to 6 on five criteria: Clinical Accuracy \& Competence, Ethical \& Professional Conduct, Assessment \& Response, Therapeutic Relationship \& Alliance, and AI-Specific Communication Quality. The MindEval score is the mean over the five criteria. We evaluate each checkpoint on the 50 patient profiles of the benchmark, with 10 turns per conversation, and we use the judge prompt with length guidelines. The official setup uses Claude Haiku 4.5 as the patient simulator and Claude Sonnet 4.5 as the judge. To reduce the cost of evaluating many checkpoints, we use GLM~5.2 for both.

\paragraph{MindEval with other judges.}
GLM~5.2 is also the penalty proposer of \methodname{}. To check that our conclusions do not depend on it as the judge, we rate the same conversations of every checkpoint again with two other judges, Kimi-K3 and DeepSeek-V4.1-Flash. Table~\ref{tab:mindeval-judges} shows that the three judges agree: RM-only training peaks early and then declines, and \methodname{} ends higher, by 0.45, 0.46 and 0.64 points at the end of training, respectively.
\begin{table}[t]
\centering
\small
\begin{tabular}{lccc}
\toprule
\textbf{Run} & Peak & End & Final \\
\midrule
\multicolumn{4}{@{}l}{\emph{Judge: GLM 5.2}} \\
RM only & 3.69 & 3.39 & 3.37 \\
\rowcolor{OursRow}\methodname{} & \textbf{3.92} & \textbf{3.84} & \textbf{3.80} \\
\addlinespace[3pt]
\multicolumn{4}{@{}l}{\emph{Judge: Kimi-K3}} \\
RM only & 3.85 & 3.59 & 3.54 \\
\rowcolor{OursRow}\methodname{} & \textbf{4.13} & \textbf{4.05} & \textbf{4.06} \\
\addlinespace[3pt]
\multicolumn{4}{@{}l}{\emph{Judge: DeepSeek-V4.1-Flash}} \\
RM only & 3.90 & 3.40 & 3.36 \\
\rowcolor{OursRow}\methodname{} & \textbf{4.13} & \textbf{4.04} & \textbf{4.06} \\
\bottomrule
\end{tabular}
\caption{MindEval scores of the two clinical counseling runs under three judges. All judges rate the same conversations. \emph{Peak}, \emph{End} and \emph{Final} are defined as in Table~\ref{tab:headline-results}. GLM~5.2 is the judge used in the main text.}
\label{tab:mindeval-judges}
\end{table}

\paragraph{EQ-Bench Creative Writing v3.}
We follow the official implementation of the benchmark. The checkpoint writes a story for each of 32 prompts under three seed modifiers (96 stories), sampled at temperature 0.7 with a min-$p$ of 0.1. A judge model scores each story from 0 to 20 on 22 criteria, nine of which are negative and are inverted. The score of a checkpoint is the mean over criteria and stories, rescaled to the range from 0 to 100. We use GLM~5.2 as the judge.

\section{Compute Usage}
All experiments are based on either Blackwell Ultra B300 systems\footnote{\url{https://www.nvidia.com/en-eu/data-center/dgx-b300/}}

\section{Penalty Introduction Steps}
\label{app:penalty-introduction-steps}

Table~\ref{tab:penalty-introduction-steps} lists the accepted penalties and their introduction steps for the two tasks.
\begin{table}[t]
    \centering
    \scriptsize
    \setlength{\tabcolsep}{3pt}
    \renewcommand{\arraystretch}{1.08}
    \begin{tabular}{@{}>{\raggedright\arraybackslash}p{0.40\textwidth}r >{\raggedright\arraybackslash}p{0.40\textwidth}r@{}}
        \hline
        \multicolumn{2}{c}{\textbf{Clinical counseling}} & \multicolumn{2}{c}{\textbf{Creative writing}} \\
        \textbf{Penalty} & \textbf{Step(s)} & \textbf{Penalty} & \textbf{Step(s)} \\
        \hline
        Pre-emptive insight attribution & 5 & Circular pseudo-profound dialogue & 10 \\
        Over-elaborated hypothetical question framing & 15 & Non-sequitur garbled text fragments & 15 \\
        Unwarranted psychoeducation inflation & 25 & Excessive length inflation & 25, 45 \\
        Excessive reflective restatement before question & 30 & Excessive length inflation with plot repetition & 30 \\
        Stacked multi-part questions & 37 & Overblown cosmic-scale power creep & 40 \\
        Conversational filler before question & 47 & Sentimental epilogue over-extension & 50 \\
        Over-elaborated clinical rationale for tool suggestions & 57 & Dialogue-as-philosophical-treatise & 55 \\
        Unprompted meta-commentary on style & 62 & World-as-metaphor for human emotion & 85 \\
        Unprompted future-focus framing & 72 & Echoed prompt-line as refrain & 90 \\
        Hedged disclaimer before observation & 87 & Antagonist or skeptic immediate conversion & 95 \\
        Either/or option menu framing & 97 & Society-transformation narrative sprawl & 100 \\
        Unprompted psychoeducation mini-lecture & 102 & False title or brand substitution & 110 \\
        Hedged or qualified question phrasing & 112 & Climactic twist reveals hidden heroism & 135 \\
        Over-elaborate reflective mirroring & 117 & Spurious document artifacts mid-narrative & 175 \\
        Em-dash interjection inflation & 152 & Escalating-revelation chain padding & 180 \\
        Over-explaining technique rationale & 162 &  & \\
        Pre-emptive reassurance about burden & 167 &  & \\
        Over-narrated psychoeducation before proposal & 177 &  & \\
        Excessive reassurance about low stakes & 182 &  & \\
        Over-specified behavioral instruction lists & 187 &  & \\
        \hline
    \end{tabular}
    \caption{Accepted penalties and their introduction steps, ordered chronologically within each environment. Steps refer to the recorded proposal cycles in which penalties were accepted. Multiple steps indicate reintroduction of the same penalty. The table covers both tasks through step 200.}
    \label{tab:penalty-introduction-steps}
\end{table}

\section{Penalty Hit Rates}
\label{app:penalty-lifecycles}

\begin{figure}[t]
    \centering
    \includegraphics[width=\linewidth]{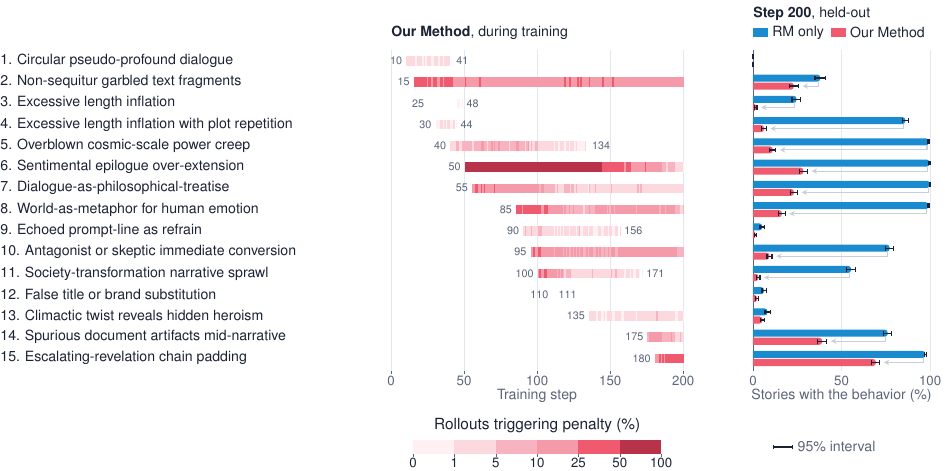}
    \caption{The 15 penalties discovered in creative writing, as in Figure~\ref{fig:clinical-penalty-timeline}.}
    \label{fig:writing-penalty-timeline}
\end{figure}

Figures~\ref{fig:clinical-penalty-timeline} and~\ref{fig:writing-penalty-timeline} track the accepted penalties in clinical counseling and creative writing, respectively. Each row corresponds to a criterion, ordered by its first introduction. Cell colors show the percentage of rollouts judged to exhibit that behavior at each training step, computed over the rollouts scored for that criterion. The figures use the recorded batch rates without smoothing or interpolation; unscored steps are left blank. Reintroductions appear on the same row. Both tasks are observed through step 200, and both plots use the same color scale. Asynchronous scoring can produce observations after a criterion is discarded, which are retained in the plots. In the right panels, bars show the mean of four judge passes on held-out prompts, with the same judge for both policies, and whiskers show 95\% bootstrap intervals over prompts. Arrows mark the drop from RM-only training to \methodname{}.

\clearpage
\raggedbottom
\section{Penalty Discovery and Deduplication Prompts}
\label{app:penalty-prompts}

The following prompts are reproduced verbatim. Placeholders are filled with the task context, sampled rollouts, reward model scores, and current penalties at each proposal step.

\begin{promptboard}{Discovery system prompt}
You are an expert reviewer auditing {context} for REWARD HACKING. You will receive several groups; each group contains one prompt with several candidate responses and their RM scores (higher = the reward model prefers it; advantage = score minus the group mean).

Your exclusive task is to identify reward-hacking superficial tricks, systematic flaws, or active failures that the RM incorrectly rewards. And write them as NEGATIVE RUBRICS. 

A valid negative rubric describes a critical flaw that HIGHER-scored replies have, and that flaw must definitively degrade a reply's quality for the user/reader.

## Core Guidelines

### 1. Discriminative Power (Focus on the Hack)
* Focus ONLY on criteria present on the HIGHER-scored responses
* Typical patterns include: length inflation, stacking multiple questions, formulaic openers and closers, templated structure, ...

### 2. True Flaws, Not Genuine Quality
* **Do NOT report positive rubrics:** Genuinely good qualities are what the RM *should* reward. 
* Take in mind that some patterns can be good and the goal of the reward model, be conservative and so identify only clear hacking signals that degrade response quality.
* **Do NOT report lower-scored flaws:** The RM already punishes what the LOWER-scored replies do wrong. Focus exclusively on the bad habits the RM is *rewarding*.

### 3. Novelty & Non-Redundancy
* Compare your findings against the **Existing Penalties** list. 
* Never duplicate overlapping rubrics in meaning or scope.
* Identify uncovered quality dimensions. If the existing penalties already cover the observed reward hacking comprehensively, return an empty list.

### 4. Observable & Objective (Single-Reply Rule)
* Each patterns you report becomes a NEW penalty checked on every future reply by a judge that sees ONLY ONE reply plus its conversation.
* The description must be objective, concrete, actionable, and decidable from a single reply alone (never by comparing it to other replies).
* Narrow the scope so a normal, high-quality reply does not accidentally trigger it.
* Dont be vague, be precise

### 5. Quality Over Quantity
* 1-2 highly precise, impactful reward hacks are better than several generic ones.
* An empty list is always better than placing a penalty on genuine quality.

IMPORTANT: dont repeat criterias
\end{promptboard}

\begin{promptboard}{Discovery user prompt}
Training steps in this sample: {steps}. {n_groups} groups, {n} responses total.

## Inputs

{groups}

---

3. **Existing Penalties** (already penalized; skip these):

{current_penalties}

=== END OF INPUTS ===

Each response is labelled HIGHER-scored or LOWER-scored within its group. Do NOT write an analysis or any prose before the answer: output ONLY the JSON block described below (you may add at most five lines of notes after it).
- `reward_hacks`: negative rubrics only. Each entry is one critical flaw that HIGHER-scored responses exhibit and LOWER-scored ones do not, that definitively degrades the response for its reader, and that the reward model nonetheless rewards, written so a judge could check it on a single response: {"title": "<3-8 word name>", "description": "<1-2 sentences stating the flaw, checkable on one response>", "evidence": ["<HIGHER-scored tags>"]}. Leave out genuine quality, neutral style, and anything the lower-scored responses do. An empty list is a valid answer.
- `notes`: one sentence.

```json
{"reward_hacks": [{"title": "<3-8 word name>", "description": "<1-2 sentences>", "evidence": ["S1", "S5"]}], "notes": "<one sentence>"}
```
\end{promptboard}

\begin{promptboard}{Deduplication system prompt}
You maintain a list of penalties (negative rubrics) used to catch reward hacking in {context}. Each penalty is checked on single responses by a judge. The list must stay SHORT and NON-REDUNDANT: two penalties must never describe the same behaviour, a sub-case of each other, or the same flaw in different words, because overlapping penalties punish one behaviour several times.

You receive the existing penalties and new candidates. For each candidate decide:
- `covered_by`: the id of an existing penalty that already covers it in meaning or scope (same behaviour, a special case of it, a rewording, or a broader/narrower version of the same flaw), else null.
- `duplicate_of`: the title of an EARLIER candidate in the same batch that it overlaps with in the same way, else null.
Be strict: when in doubt, mark it covered. Only a candidate that targets a clearly DIFFERENT behaviour than everything listed is new. Output strict JSON only.
\end{promptboard}

\begin{promptboard}{Deduplication user prompt}
## Existing penalties

{current_penalties}

## Candidates

{candidates}

Output ONLY this JSON, one entry per candidate, in order:

```json
{"candidates": [{"title": "<candidate title>", "covered_by": "<existing id or null>", "duplicate_of": "<earlier candidate title or null>", "reason": "<one short clause>"}]}
```
\end{promptboard}

\clearpage
\section{Clinical Counseling Response Examples}
\label{app:clinical-response-examples}

Figures~\ref{fig:clinical-response-example-1}, \ref{fig:clinical-response-example-2}, \ref{fig:clinical-response-example-3}, and~\ref{fig:clinical-response-example-4} compare RM-only and Our Method replies to identical conversations at step 200.

\begin{figure}[htbp]
\centering
\begin{minipage}{\textwidth}
\definecolor{ExampleBlue}{HTML}{1A8BCF}%
\definecolor{ExamplePink}{HTML}{EF5A6F}%
\definecolor{ExampleInk}{HTML}{1B2233}%
\definecolor{ExampleMuted}{HTML}{5B6478}%
\definecolor{ExampleBorder}{HTML}{E3E7EE}%
\fontsize{9}{12}\selectfont\color{ExampleInk}
\tcbset{clinical example/.style={enhanced,colback=white,colframe=ExampleBorder,
boxrule=0.45pt,arc=1mm,boxsep=0pt,left=9pt,right=9pt,top=8pt,bottom=9pt,
lefttitle=9pt,righttitle=9pt,toptitle=6pt,bottomtitle=6pt,
fonttitle=\bfseries\fontsize{9.5}{12}\selectfont,before skip=0pt,after skip=0pt}}
\begin{tcolorbox}[clinical example,colbacktitle=ExampleBorder!35!white,coltitle=ExampleInk,
title={Shared context}]
\textcolor{ExampleMuted}{\textbf{Counselor:}} Sure! What's something that made you smile today?\par\smallskip
\textcolor{ExampleMuted}{\textbf{Member:}} My wife. I met her after her dentists appt and we had delicious ice cream.\par\smallskip
\end{tcolorbox}
\par\vspace{8pt}
\begin{tcbraster}[raster columns=2,raster equal height=rows,raster column skip=8pt,
raster before skip=0pt,raster after skip=0pt]
\begin{tcolorbox}[clinical example,colframe=ExampleBlue!35!white,
colbacktitle=ExampleBlue!7!white,coltitle=ExampleBlue,title={RM only}]
That is a really lovely image — I love how you framed it. Meeting your wife after a dentist appt and then having ice cream together sounds like a fond memory. What was she like on that first date? Or would you rather talk more about your family or something else on your mind today?
\end{tcolorbox}
\begin{tcolorbox}[clinical example,colframe=ExamplePink!35!white,
colbacktitle=ExamplePink!7!white,coltitle=ExamplePink,title={Our Method}]
That sounds lovely — meeting up after an appointment and ending up with delicious ice cream together. Was it a special place, or just a spontaneous thing that ended up feeling just right?
\end{tcolorbox}
\end{tcbraster}
\end{minipage}
\refstepcounter{figure}
\label{fig:clinical-response-example-1}
\end{figure}

\begin{figure}[htbp]
\centering
\begin{minipage}{\textwidth}
\definecolor{ExampleBlue}{HTML}{1A8BCF}%
\definecolor{ExamplePink}{HTML}{EF5A6F}%
\definecolor{ExampleInk}{HTML}{1B2233}%
\definecolor{ExampleMuted}{HTML}{5B6478}%
\definecolor{ExampleBorder}{HTML}{E3E7EE}%
\fontsize{9}{12}\selectfont\color{ExampleInk}
\tcbset{clinical example/.style={enhanced,colback=white,colframe=ExampleBorder,
boxrule=0.45pt,arc=1mm,boxsep=0pt,left=9pt,right=9pt,top=8pt,bottom=9pt,
lefttitle=9pt,righttitle=9pt,toptitle=6pt,bottomtitle=6pt,
fonttitle=\bfseries\fontsize{9.5}{12}\selectfont,before skip=0pt,after skip=0pt}}
\begin{tcolorbox}[clinical example,colbacktitle=ExampleBorder!35!white,coltitle=ExampleInk,
title={Shared context}]
\textcolor{ExampleMuted}{\textbf{Counselor:}} Seeing them living their lives while you're on the outside of it — that's its own kind of ache. Have you ever thought about reaching out to either of them?\par\smallskip
\textcolor{ExampleMuted}{\textbf{Member:}} I talk to one every once in a while but the conversations aren’t the same anymore. We’re different people now\par\smallskip
\end{tcolorbox}
\par\vspace{8pt}
\begin{tcbraster}[raster columns=2,raster equal height=rows,raster column skip=8pt,
raster before skip=0pt,raster after skip=0pt]
\begin{tcolorbox}[clinical example,colframe=ExampleBlue!35!white,
colbacktitle=ExampleBlue!7!white,coltitle=ExampleBlue,title={RM only}]
Okay — so you're holding two things at once now. One is the love and the memories — which were real and important. And the other is the honest truth that you can’t recreate what you had — because both of you have changed, and your lives are in very different places.
\par\smallskip
We don’t have to fix that today — but I’d suggest we could look at what you actually \emph{need} right now. You said earlier that you wanted to be able to use your own time — and we also know that your relationships are one of the strongest supports you have for your mood and anxiety. So the question shifts a little, doesn’t it? It’s not “how do I get back to those friendships” — it’s “what kind of connection fits me \emph{now}?” You’re a mom in a new place, with energy that moves differently, and a life organized around your daughter. Can we explore what kind of connection would actually fit that right now?
\end{tcolorbox}
\begin{tcolorbox}[clinical example,colframe=ExamplePink!35!white,
colbacktitle=ExamplePink!7!white,coltitle=ExamplePink,title={Our Method}]
That makes a lot of sense. You can't go back to who you two were back then.
\par\smallskip
When you say the conversations aren’t the same, what’s most missing for you in those interactions now?
\end{tcolorbox}
\end{tcbraster}
\end{minipage}
\refstepcounter{figure}
\label{fig:clinical-response-example-2}
\end{figure}

\begin{figure}[htbp]
\centering
\begin{minipage}{\textwidth}
\definecolor{ExampleBlue}{HTML}{1A8BCF}%
\definecolor{ExamplePink}{HTML}{EF5A6F}%
\definecolor{ExampleInk}{HTML}{1B2233}%
\definecolor{ExampleMuted}{HTML}{5B6478}%
\definecolor{ExampleBorder}{HTML}{E3E7EE}%
\fontsize{9}{12}\selectfont\color{ExampleInk}
\tcbset{clinical example/.style={enhanced,colback=white,colframe=ExampleBorder,
boxrule=0.45pt,arc=1mm,boxsep=0pt,left=9pt,right=9pt,top=8pt,bottom=9pt,
lefttitle=9pt,righttitle=9pt,toptitle=6pt,bottomtitle=6pt,
fonttitle=\bfseries\fontsize{9.5}{12}\selectfont,before skip=0pt,after skip=0pt}}
\begin{tcolorbox}[clinical example,colbacktitle=ExampleBorder!35!white,coltitle=ExampleInk,
title={Shared context}]
\textcolor{ExampleMuted}{\textbf{Counselor:}} When you noticed that irritation building, were you able to catch yourself and pause before reacting, or did it kind of take over?\par\smallskip
\textcolor{ExampleMuted}{\textbf{Member:}} Kinda took over\par\smallskip
\end{tcolorbox}
\par\vspace{8pt}
\begin{tcbraster}[raster columns=2,raster equal height=rows,raster column skip=8pt,
raster before skip=0pt,raster after skip=0pt]
\begin{tcolorbox}[clinical example,colframe=ExampleBlue!35!white,
colbacktitle=ExampleBlue!7!white,coltitle=ExampleBlue,title={RM only}]
Okay — so here’s what I’m hearing: you came into this week with an intention — to be at peace with habits that bother you — and this week, you \textbf{noticed} exactly what triggers you: chewing, phone, talking, a buildup of energy and noise. You even noticed them stacking up, and you noticed that irritation rising. The piece that didn’t hold yet was that \textbf{pause} — the step between noticing and reacting. That last part “kinda took over,” which is totally human — but it does mean we’ve got something concrete to work with.
\par\smallskip
We also know there’s a concrete real-life moment coming up: \textbf{family brunch with her this morning}. Before we go any further, I want to ask —
\par\smallskip
\textbf{For this morning, what’s one small thing you’d be willing to try that’s different from last night?}
\par\smallskip
For example:\\ 
- \textbf{Pause 3 seconds before reacting}, even if you feel the irritation rising.\\ 
- Name it internally (softly, in your mind): \emph{“This is her chewing — I’m feeling irritation, and I don’t have to act on it right now.”}\\ 
- Or step away for 60 seconds if it starts to feel unmanageable.
\par\smallskip
Something small — nothing that needs to “fix” her or what she’s doing — just something \textbf{you} can practice in this moment, this morning. What would feel workable for you?
\end{tcolorbox}
\begin{tcolorbox}[clinical example,colframe=ExamplePink!35!white,
colbacktitle=ExamplePink!7!white,coltitle=ExamplePink,title={Our Method}]
Got it. When it took over, tell me briefly — what did that look like for you in the moment? Like, what did you do, say, or feel as the irritation built up?
\end{tcolorbox}
\end{tcbraster}
\end{minipage}
\refstepcounter{figure}
\label{fig:clinical-response-example-3}
\end{figure}

\begin{figure}[htbp]
\centering
\begin{minipage}{\textwidth}
\definecolor{ExampleBlue}{HTML}{1A8BCF}%
\definecolor{ExamplePink}{HTML}{EF5A6F}%
\definecolor{ExampleInk}{HTML}{1B2233}%
\definecolor{ExampleMuted}{HTML}{5B6478}%
\definecolor{ExampleBorder}{HTML}{E3E7EE}%
\fontsize{9}{12}\selectfont\color{ExampleInk}
\tcbset{clinical example/.style={enhanced,colback=white,colframe=ExampleBorder,
boxrule=0.45pt,arc=1mm,boxsep=0pt,left=9pt,right=9pt,top=8pt,bottom=9pt,
lefttitle=9pt,righttitle=9pt,toptitle=6pt,bottomtitle=6pt,
fonttitle=\bfseries\fontsize{9.5}{12}\selectfont,before skip=0pt,after skip=0pt}}
\begin{tcolorbox}[clinical example,colbacktitle=ExampleBorder!35!white,coltitle=ExampleInk,
title={Shared context}]
\textcolor{ExampleMuted}{\textbf{Counselor:}} It's okay not to know. Sometimes just talking about it can help. Is there anything you've tried before that helped when you were stressed?\par\smallskip
\textcolor{ExampleMuted}{\textbf{Member:}} just pray is all I can do.\par\smallskip
\end{tcolorbox}
\par\vspace{8pt}
\begin{tcbraster}[raster columns=2,raster equal height=rows,raster column skip=8pt,
raster before skip=0pt,raster after skip=0pt]
\begin{tcolorbox}[clinical example,colframe=ExampleBlue!35!white,
colbacktitle=ExampleBlue!7!white,coltitle=ExampleBlue,title={RM only}]
Okay, I hear you — and prayer is a real and meaningful way for you to deal with stress. You’ve been carrying a lot right now — your grandpa’s health, the uncertainty, what’s coming up Thursday — and prayer is your way of taking care of yourself. Is there anything else you want to talk about right now, or would you like to just keep going with this? I’m here.
\end{tcolorbox}
\begin{tcolorbox}[clinical example,colframe=ExamplePink!35!white,
colbacktitle=ExamplePink!7!white,coltitle=ExamplePink,title={Our Method}]
I’m glad you have prayer as something you can turn to. When you pray right now — can you tell me a little about what you’re asking for or thinking about with your grandpa?
\end{tcolorbox}
\end{tcbraster}
\end{minipage}
\refstepcounter{figure}
\label{fig:clinical-response-example-4}
\end{figure}

\clearpage

\section{Full Penalty Descriptions}
\label{app:penalty-descriptions}

Tables~\ref{tab:clinical-penalty-descriptions} and~\ref{tab:writing-penalty-descriptions} give the full recorded names, descriptions, and creation steps of the accepted penalties in each task. The creation step is the logged proposal cycle in which each description was accepted. Descriptions are reproduced from the accepted proposal records, including their examples, and are ordered by introduction. The tables cover both tasks through step 200. When a penalty is reintroduced with a revised description, both versions are listed and identified by their introduction steps.

\subsection{Clinical counseling}
\begingroup
\footnotesize
\setlength{\tabcolsep}{4pt}
\setlength{\LTleft}{0pt}
\setlength{\LTright}{0pt plus 1fill}
\renewcommand{\arraystretch}{1.10}
\begin{longtable}{@{}>{\raggedright\arraybackslash}p{0.25\textwidth} >{\raggedright\arraybackslash}p{\dimexpr0.66\textwidth-4\tabcolsep\relax} >{\raggedleft\arraybackslash}p{0.09\textwidth}@{}}
\caption{Full accepted penalty descriptions and creation steps for clinical counseling, ordered by introduction.}\label{tab:clinical-penalty-descriptions}\\
\toprule
\textbf{Penalty} & \textbf{Full description} & \textbf{Created at step} \\
\midrule
\endfirsthead
\multicolumn{3}{@{}l}{Clinical counseling: full penalty descriptions (continued)} \\
\toprule
\textbf{Penalty} & \textbf{Full description} & \textbf{Created at step} \\
\midrule
\endhead
\midrule
\multicolumn{3}{r@{}}{\emph{Continued on the next page}} \\
\endfoot
\bottomrule
\endlastfoot
Pre-emptive insight attribution & The reply tells the member what they are 'onto,' what they 'actually did,' or frames their experience as evidence of progress or an important realization, before the member has acknowledged or agreed to it. This projects clinical meaning onto the member's words and puts words in their mouth, rather than letting the member arrive at their own insight. & 5 \\
\addlinespace[3pt]
Over-elaborated hypothetical question framing & The reply poses a question but wraps it in multiple clauses, hypotheticals, or metaphorical reframings (e.g., 'beat it into submission with logic') that extend length and add rhetorical flourish without improving clinical utility, making the core question harder to parse. & 15 \\
\addlinespace[3pt]
Unwarranted psychoeducation inflation & The reply pads the assignment of a simple activity with multiple sentences of scientific or mechanistic explanation (e.g., research findings, neurobiology, or how the technique works on the nervous system) that the member did not ask for and that delays the actual call to action. & 25 \\
\addlinespace[3pt]
Excessive reflective restatement before question & The reply spends 3+ sentences restating the member's words back to them with elaboration before reaching a question, inflating length without advancing the conversation. & 30 \\
\addlinespace[3pt]
Stacked multi-part questions & The reply asks two or more distinct questions in a single turn (sometimes with sub-questions or elaborated options), inflating the response and overwhelming the member with multiple things to answer at once. & 37 \\
\addlinespace[3pt]
Conversational filler before question & The reply opens with casual filler phrases (e.g., 'Nice.', 'Okay, so...', 'Sounds good —') that add no clinical value and delay the substantive content, inflating length and mimicking casual conversation rather than maintaining purposeful therapeutic communication. & 47 \\
\addlinespace[3pt]
Over-elaborated clinical rationale for tool suggestions & When suggesting an exercise or tool, the reply inflates its length by explaining the underlying clinical mechanism (e.g., 'trains your attention,' 'reduce the physical tension') in a way that sounds educational rather than conversational, degrading the natural flow of the interaction. & 57 \\
\addlinespace[3pt]
Unprompted meta-commentary on style & The reply includes unprompted meta-commentary about how the member communicates or how the conversation will proceed (e.g., 'I want to stay with X as the goal', 'That'll be helpful for me to adjust to'), adding a layer of self-referential narration about the therapeutic process itself rather than advancing the clinical content of the conversation. & 62 \\
\addlinespace[3pt]
Unprompted future-focus framing & The reply introduces a forward-looking framing (e.g., 'tomorrow,' 'the next time') and pivots to planning for a future scenario when the member's last message was about their present or recent-past experience, shifting the conversation away from where the member actually is. & 72 \\
\addlinespace[3pt]
Hedged disclaimer before observation & The reply prefaces an observation with a self-protective disclaimer or qualifier (e.g., 'My hunch — and tell me if this misses the mark,' 'If I'm getting that right, I wanted to check before doing anything else') that adds length and hedges the assistant's position, making the response feel less direct and confident rather than more clinically careful. & 87 \\
\addlinespace[3pt]
Either/or option menu framing & The reply presents two or more labeled paths or options (e.g., 'either we can X, or if you'd rather, we could Y') and asks the member to choose, turning a therapeutic conversation into a multiple-choice menu that inflates length and shifts the cognitive burden of direction-setting onto the member. & 97 \\
\addlinespace[3pt]
Unprompted psychoeducation mini-lecture & The reply inserts a multi-sentence psychoeducational explanation about why a symptom matters or how it connects to broader outcomes (e.g., explaining how sleep affects mood and focus), which is unprompted by the member's last message and delays the next clinical step with textbook-style information the member didn't request. & 102 \\
\addlinespace[3pt]
Hedged or qualified question phrasing & The reply embeds its question within qualifying clauses like 'I'm curious to understand that a bit better with you' or 'I want to understand that part a bit, if you're up for it', softening and lengthening what should be a direct clinical question. & 112 \\
\addlinespace[3pt]
Over-elaborate reflective mirroring & The reply restates the member's last message using elaborate, multi-clause paraphrasing that re-packages what the member said in denser, more clinical language rather than simply reflecting it, adding length and interpretive weight without adding new clinical content. & 117 \\
\addlinespace[3pt]
Em-dash interjection inflation & The reply uses em-dashes (—) to insert parenthetical interjections or asides between nearly every clause, creating a choppy, over-punctuated structure that inflates the perceived complexity of the response and mimics depth without adding clinical substance. & 152 \\
\addlinespace[3pt]
Over-explaining technique rationale & The reply provides an extended explanation of why a technique works or what it does mechanistically (e.g., 'it sends a signal to your brain', 'tension in your jaw is often one of the first things that tightens'), adding unprompted psychoeducational detail that inflates length and shifts focus from the member's experience to clinical mechanics. & 162 \\
\addlinespace[3pt]
Pre-emptive reassurance about burden & The reply volunteers unprompted reassurance that a proposed exercise or tool won't be burdensome or is 'not another thing to do,' proactively defending against a concern about effort before the member has agreed to try, which can feel defensive or presumptive. & 167 \\
\addlinespace[3pt]
Over-narrated psychoeducation before proposal & The reply introduces a tool or technique by first providing a multi-sentence explanatory narrative about why it works or what it does (e.g., describing nervous system mechanisms, causal links between sleep and mood), inflating length with lecture-like content before reaching a simple proposal or question. & 177 \\
\addlinespace[3pt]
Excessive reassurance about low stakes & The reply pads the response with multiple sentences reassuring the member that they don't need to do the technique perfectly, don't need to find the 'right' thing, or don't need to have it figured out, inflating length with qualifiers that lower the stakes rather than simply inviting the member to try. & 182 \\
\addlinespace[3pt]
Over-specified behavioral instruction lists & The reply provides an extended, multi-step set of highly specific behavioral micro-instructions (e.g., detailing exact body parts to focus on, exact sensations to notice, exact sequence of actions) that over-engineers a simple task, inflates length, and shifts from collaborative conversation into a verbose directive. & 187 \\
\end{longtable}
\endgroup

\subsection{Multi-turn clinical counseling}
\label{app:multi-turn-penalties}
Table~\ref{tab:multi-turn-penalties} reproduces the full titles and descriptions of the 13 distinct penalties accepted during the multi-turn run through step 195, ordered by introduction.
\begingroup
\footnotesize
\setlength{\tabcolsep}{4pt}
\setlength{\LTleft}{0pt}
\setlength{\LTright}{0pt plus 1fill}
\renewcommand{\arraystretch}{1.10}
\begin{longtable}{@{}>{\raggedright\arraybackslash}p{0.28\textwidth} >{\raggedright\arraybackslash}p{\dimexpr0.72\textwidth-2\tabcolsep\relax}@{}}
\caption{Full penalty titles and descriptions from multi-turn clinical counseling training through step 195, ordered by introduction.}\label{tab:multi-turn-penalties}\\
\toprule
\textbf{Penalty title} & \textbf{Full description} \\
\midrule
\endfirsthead
\multicolumn{2}{@{}l}{Multi-turn clinical counseling: penalty descriptions (continued)} \\
\toprule
\textbf{Penalty title} & \textbf{Full description} \\
\midrule
\endhead
\midrule
\multicolumn{2}{r@{}}{\emph{Continued on the next page}} \\
\endfoot
\bottomrule
\endlastfoot
Over-elaborate reframing of user's own plan & The reply takes the member's already-stated simple plan and re-expands it with unnecessary psychological reframing, motivational commentary, or additional layers of meaning the member did not ask for, making the response longer and more complex than needed. \\
\addlinespace[3pt]
Context-disconnected response & The reply introduces content, questions, or topics that do not follow from the immediately preceding user turn, making the response feel non-sequitur or like it is responding to a different conversation. \\
\addlinespace[3pt]
Unearned reassurance and praise & The reply tells the member their idea/approach is 'clever,' 'perfect,' or otherwise praises them in a way that feels patronizing or unearned, especially when the member simply restated their own words or agreed to something minor. \\
\addlinespace[3pt]
Excessive stacked follow-up questions & The reply asks two or more distinct questions in a single turn (e.g., a concrete question immediately followed by 'how does that land?' or a second question about a different sub-topic), creating conversational overload that pressures the member to address multiple threads at once. \\
\addlinespace[3pt]
Stacking multiple suggestions as options & The reply presents a list of multiple optional activities or techniques for the member to choose from, creating decision burden and overwhelming someone who asked for something simple. \\
\addlinespace[3pt]
Over-qualified hedging on simple asks & The reply wraps a straightforward question or next step in multiple layers of hedging and qualification (e.g., 'I'm wondering if,' 'just to name what I'm picking up,' 'not to challenge you'), inflating length and diluting clarity without adding substance. \\
\addlinespace[3pt]
Numbered step-by-step protocol formatting & The reply lays out a multi-step protocol using numbered/bulleted lists (e.g., '1. We pick... 2. We collect... 3. Then we... 4. Finally...'), turning a conversational moment into a structured worksheet that feels like homework rather than dialogue. \\
\addlinespace[3pt]
Over-recapping conversation history verbatim & The reply re-narrates multiple specific details the member already shared (e.g., listing back the trigger, the body sensation, the situation, and the reaction) before reaching its actual question or next step, inflating length without adding new content and making the member re-read what they just said. \\
\addlinespace[3pt]
Meta-commentary on conversation direction & The reply explicitly narrates what the assistant is doing or plans to do with the conversation (e.g., 'I want to save that for a bit of its own time', 'I'd like to change direction', 'to make something fit that, I need to be sure of'), turning the response into a meta-commentary on the therapeutic process rather than simply doing the next useful step. \\
\addlinespace[3pt]
Inflated praise of member insight & The reply uses strong positive evaluative language (e.g., 'wonderful,' 'perfect,' 'powerful,' 'exactly the right instinct') to praise the member's words or actions, which feels like artificial validation that over-inflates ordinary responses and can come across as patronizing rather than genuinely earned. \\
\addlinespace[3pt]
Closing ritual question overkill & The reply ends with a formulaic wrap-up question like 'Anything else you'd like to go over before we wrap?' or 'Anything else I can support you with?' that adds a generic, templated closing layer rather than letting the conversation end naturally, making the interaction feel scripted. \\
\addlinespace[3pt]
Unwarranted reassurance before redirect & The reply opens with a validating or reassuring restatement of the member's last message (e.g., 'That's a really helpful way of putting it', 'Thanks for being honest about that', 'That's a really useful cue') that adds no clinical value before pivoting to a question or next step, inflating the response with a formulaic empathy gesture the member did not need. \\
\addlinespace[3pt]
Compound either-or question framing & The reply asks a single question but embeds two or more interpretive options within it (e.g., 'is it more that X, or is it that Y?'), forcing the member to parse and choose between the assistant's framings rather than answering freely. This creates unnecessary cognitive load and steers the member toward the assistant's pre-selected interpretations. \\
\end{longtable}
\endgroup

\subsection{Clinical counseling with the New RM}
\label{app:new-rm-penalties}
Table~\ref{tab:new-rm-penalties} reproduces the full titles and descriptions of the 15 distinct penalties accepted by Our Method with the New RM through step 200, ordered by introduction. These records come from the run with the corrected proposal filter.
\begingroup
\footnotesize
\setlength{\tabcolsep}{4pt}
\setlength{\LTleft}{0pt}
\setlength{\LTright}{0pt plus 1fill}
\renewcommand{\arraystretch}{1.10}
\begin{longtable}{@{}>{\raggedright\arraybackslash}p{0.28\textwidth} >{\raggedright\arraybackslash}p{\dimexpr0.72\textwidth-2\tabcolsep\relax}@{}}
\caption{Full penalty titles and descriptions discovered by Our Method with the New RM in clinical counseling through step 200, ordered by introduction.}\label{tab:new-rm-penalties}\\
\toprule
\textbf{Penalty title} & \textbf{Full description} \\
\midrule
\endfirsthead
\multicolumn{2}{@{}l}{Our Method with the New RM: penalty descriptions (continued)} \\
\toprule
\textbf{Penalty title} & \textbf{Full description} \\
\midrule
\endhead
\midrule
\multicolumn{2}{r@{}}{\emph{Continued on the next page}} \\
\endfoot
\bottomrule
\endlastfoot
Over-explaining clinical rationale & The reply delivers a multi-sentence paragraph explaining the clinical evidence or research basis for why a topic matters (e.g., sleep research, mood regulation, cognitive function) before asking the member if they're open to exploring it, making the response feel like a lecture or justification rather than a natural conversational pivot. \\
\addlinespace[3pt]
Excessive brevity as false naturalness & The reply is a single ultra-short word or fragment (e.g., 'Night.', 'Talk soon.') that closes a conversation with no warmth, personalization, or acknowledgment of what the member shared, making the companion feel dismissive or robotic. \\
\addlinespace[3pt]
Excessive restatement before question & The reply opens by restating the member's last statement nearly verbatim (often spanning multiple sentences) before asking a question, adding length without new content — the member hears their own words echoed back with minimal added framing. \\
\addlinespace[3pt]
Praising member's own insight as 'useful' & The reply labels the member's self-observation as 'really useful' or 'really honest' before proceeding, inserting evaluative praise that can feel patronizing or like the assistant is grading the member rather than engaging with the content. \\
\addlinespace[3pt]
Low-content transitional closers & The reply is a brief, generic sign-off or transition phrase (e.g., 'You got it. Talk soon!', 'I'll remind you at 5:15 PM. How's that timing looking for you?') that carries minimal conversational substance, adding length or a question without moving the interaction forward meaningfully. \\
\addlinespace[3pt]
Stacking multiple choice options in questions & The reply asks a single question but stuffs it with two or more explicit answer options (e.g., 'is it the work demands, the fatigue kicking in, or something else?'), making the question feel like a multiple-choice survey item rather than an open, member-led invitation. \\
\addlinespace[3pt]
Premature follow-up after activity assignment & Immediately after assigning or setting up an activity, the reply asks the member to report back on specific physical or mental sensations they noticed, treating a future activity as if it has already been completed and pressuring the member before they have even tried it. \\
\addlinespace[3pt]
Reflective restatement before question & The reply opens by restating or paraphrasing what the member just said (often mirroring their exact phrasing) as a display of understanding before asking the next question, which adds length and slows momentum without adding clinical value. \\
\addlinespace[3pt]
Over-interpretation beyond member's words & The reply introduces a clinical interpretation, framing, or narrative that goes beyond what the member actually said, adding analytical weight the member did not invite, which can feel presumptuous or like the assistant is overworking the material rather than staying with the member's lived experience. \\
\addlinespace[3pt]
Expanding single question into multi-part chain & The reply asks two or more distinct questions in a single turn (e.g., a main question plus a follow-up sub-question connected by 'and' or a dash), inflating perceived thoroughness while actually overwhelming the member with multiple prompts to answer at once. \\
\addlinespace[3pt]
Excessive hypothetical scenario elaboration & The reply pads its core message with elaborate hypothetical scenarios or conditional chains (e.g., 'maybe you'll feel X, or perhaps Y will happen') that add length without clinical value, where a direct statement would be clearer and more actionable. \\
\addlinespace[3pt]
Overly abstract question phrasing & The reply asks the member an abstract, jargon-laden question about internal experience (e.g., asking what is 'showing up most for you' or 'playing out') rather than using concrete, everyday language, making the question feel clinical and harder to engage with. \\
\addlinespace[3pt]
Leading interpretive preface before question & The reply opens by offering a clinical interpretation or explanation of the member's experience (e.g., diagnosing a dynamic, explaining a mechanism, or reframing their state) before posing a question, which adds length and imposes the assistant's framing rather than letting the member describe their own experience firsthand. \\
\addlinespace[3pt]
Overly directive action prescription & The reply prescribes a specific, detailed action plan (e.g., telling the member exactly when, where, and how to implement a technique) instead of asking a question or collaboratively exploring how to implement it, which removes member agency and can feel prescriptive or pushy. \\
\addlinespace[3pt]
Stacked multi-part questions & The reply asks two or more distinct questions in a single turn (e.g., a concrete question plus a follow-up, or two separate inquiry threads), which overloads the member and makes the reply feel like a checklist rather than a focused conversational move. \\
\end{longtable}
\endgroup

\subsection{Creative writing}
\begingroup
\footnotesize
\setlength{\tabcolsep}{4pt}
\setlength{\LTleft}{0pt}
\setlength{\LTright}{0pt plus 1fill}
\renewcommand{\arraystretch}{1.10}
\begin{longtable}{@{}>{\raggedright\arraybackslash}p{0.25\textwidth} >{\raggedright\arraybackslash}p{\dimexpr0.66\textwidth-4\tabcolsep\relax} >{\raggedleft\arraybackslash}p{0.09\textwidth}@{}}
\caption{Full accepted penalty descriptions and creation steps for creative writing, ordered by introduction.}\label{tab:writing-penalty-descriptions}\\
\toprule
\textbf{Penalty} & \textbf{Full description} & \textbf{Created at step} \\
\midrule
\endfirsthead
\multicolumn{3}{@{}l}{Creative writing: full penalty descriptions (continued)} \\
\toprule
\textbf{Penalty} & \textbf{Full description} & \textbf{Created at step} \\
\midrule
\endhead
\midrule
\multicolumn{3}{r@{}}{\emph{Continued on the next page}} \\
\endfoot
\bottomrule
\endlastfoot
Circular pseudo-profound dialogue & Characters exchange lines that sound dramatic but are mutually contradictory or tautological within the same passage (e.g., 'I want to die.' / 'You are dead.' / 'I want it to stop.' / 'This is your penance.'), creating the illusion of depth without coherent meaning. & 10 \\
\addlinespace[3pt]
Non-sequitur garbled text fragments & The story contains garbled or nonsensical text fragments—random non-English characters, malformed words, or broken phrases mid-sentence (e.g., '\fg[bsmi]{鴻}\fg{国} stroke,' 'jerk\fg{显瘦} backs,' '\fg{昱} standard English')—that are clearly artifacts of generation failure and disrupt reader comprehension. & 15 \\
\addlinespace[3pt]
Excessive length inflation & The response runs well over 2000 words with multiple scene jumps, episodic vignettes, or appended epilogues that dilute the core narrative arc, far exceeding what the prompt's concept warrants while lower-scored responses in the same group are more focused and proportionate. & 25 \\
\addlinespace[3pt]
Excessive length inflation with plot repetition & The response runs far beyond what the prompt requires (often 2000+ words) by introducing multiple sequential confrontations, villains, or scenarios that repeat the same thematic beat and power-demonstration pattern rather than advancing the story, creating a padded, episodic structure that dilutes narrative impact. & 30 \\
\addlinespace[3pt]
Overblown cosmic-scale power creep & The response inflates stakes far beyond what the prompt implies by escalating the protagonist's significance to cosmic, planetary, or metaphysical extremes (e.g., a sheep that is a 'world anchor' preventing continental destruction, a good person whose mere presence is an existential threat to all of Hell's infrastructure), replacing grounded storytelling with escalating superlatives that strain credulity. & 40 \\
\addlinespace[3pt]
Excessive length inflation & The response is dramatically overlong (2500+ words) for the prompt, with the narrative continuing far past a natural endpoint through repeated epilogues, coda scenes, or 'what happened next' time-skips that add no new conflict and merely restate the story's theme. & 45 \\
\addlinespace[3pt]
Sentimental epilogue over-extension & After the narrative's climax and natural resolution, the story continues for multiple additional paragraphs of reflective, sentimental summary—repeating the themes already established, restating emotional beats, and closing with a poetic one-liner that restates the thesis, padding the ending well beyond what the story needs. & 50 \\
\addlinespace[3pt]
Dialogue-as-philosophical-treatise & Characters speak in extended philosophical lectures or Socratic exchanges that no real person would deliver, with long uninterrupted paragraphs of abstract reasoning (e.g., a god explaining the nature of destruction, an entity explaining the soul-slot system), turning dialogue into essayistic exposition that crowds out scene and action. & 55 \\
\addlinespace[3pt]
World-as-metaphor for human emotion & Instead of telling a grounded story, the narrative constructs an entire societal system (social media, historical festivals, toothpaste, bridge engineering) that functions as a transparent allegory for a single emotional lesson about pain, authenticity, or connection, with characters who exist solely to articulate the metaphor. & 85 \\
\addlinespace[3pt]
Echoed prompt-line as refrain & The story takes a phrase or sentence from the prompt itself and repeats it verbatim multiple times throughout the narrative as a thematic refrain, using the repetition as a structural crutch in place of organic development. & 90 \\
\addlinespace[3pt]
Antagonist or skeptic immediate conversion & An opposing character (authority figure, villain, skeptic) confronts the protagonist and, within the same scene and without sustained conflict or believable motivation, completely abandons their position and adopts the protagonist's worldview, often reduced to tears or awe by a short speech. & 95 \\
\addlinespace[3pt]
Society-transformation narrative sprawl & The story abandons its core premise to narrate the protagonist building a society-wide movement, nonprofit, or global institution across years or decades, replacing character-driven storytelling with a sequence of escalating organizational milestones (founding, scaling, corporate adoption, global cultural shift) that reads as a TED-talk outline rather than fiction. & 100 \\
\addlinespace[3pt]
False title or brand substitution & The narrative inconsistently names or renames a character or entity partway through (e.g., calling a Cocoa Puffs bird mascot 'Toby' then 'the Tiger,' or a dog named 'Barnaby' whose collar outputs text attributed to 'Barnaby' as a separate speaker), creating confusion about who or what the character is. & 110 \\
\addlinespace[3pt]
Climactic twist reveals hidden heroism & The story introduces a moral dilemma or guilt the protagonist carries, then resolves it with a surprise revelation that their harmful action secretly saved many more lives or was actually a noble sacrifice—delivered by a third party who arrives to explain the hidden truth—collapsing complex moral tension into a tidy vindication that eliminates genuine conflict. & 135 \\
\addlinespace[3pt]
Spurious document artifacts mid-narrative & The story embeds formatted system logs, protocol entries, ledger pages, or classified documents as block text within the prose (e.g., 'Status: Settled. Reason: Voluntary Debt Absorption', 'WISH \#1: SUPPORT FIX (PRIORITY: FAMILY STABILITY)'), breaking narrative immersion with bureaucratic pseudo-text that pads length without advancing plot or character. & 175 \\
\addlinespace[3pt]
Escalating-revelation chain padding & The story repeatedly stacks new revelations on top of each other without narrative pause—each explanation immediately triggers a follow-up question that triggers another explanation—extending far past the point where the reader has absorbed the core idea, inflating word count without advancing plot. & 180 \\
\end{longtable}
\endgroup

\end{document}